\documentclass{article}

\usepackage{microtype}
\usepackage{graphicx}
\usepackage{caption}
\usepackage{subcaption}
\usepackage{booktabs} 

\usepackage{url}

\usepackage{breakurl}
\usepackage[breaklinks]{hyperref}

\usepackage[accepted]{arxiv}

\usepackage{amsmath}
\usepackage{amssymb}
\usepackage{mathtools}
\usepackage{amsthm}

\usepackage[capitalize,noabbrev]{cleveref}

\theoremstyle{plain}

\theoremstyle{definition}

\theoremstyle{remark}

\usepackage[textsize=tiny]{todonotes}

\icmltitlerunning{Tokenization counts: the impact of tokenization on arithmetic in frontier LLMs}

\begin{document}

\twocolumn[
\icmltitle{Tokenization counts: the impact of tokenization on arithmetic in frontier LLMs}

\icmlsetsymbol{equal}{*}

\begin{icmlauthorlist}
\icmlauthor{Aaditya K. Singh}{gatsby}
\icmlauthor{DJ Strouse}{gdm}
\end{icmlauthorlist}

\icmlaffiliation{gatsby}{Gatsby Computational Neuroscience Unit, University College London}
\icmlaffiliation{gdm}{Google DeepMind}

\icmlcorrespondingauthor{Aaditya Singh}{aaditya.singh.21@ucl.ac.uk}

\icmlkeywords{LLMs, tokenizers, tokenization, transformers, math, reasoning, Machine Learning, ICML}

\vskip 0.3in
]

\printAffiliationsAndNotice{}

\begin{abstract}
Tokenization, the division of input text into input tokens, is an often overlooked aspect of the large language model (LLM) pipeline and could be the source of useful or harmful inductive biases. Historically, LLMs have relied on byte pair encoding, without care to specific input domains. With the increased use of LLMs for reasoning, various number-specific tokenization schemes have been adopted, with popular models like LLaMa and PaLM opting for single-digit tokenization while GPT-3.5 and GPT-4 have separate tokens for each 1-, 2-, and 3-digit numbers. In this work, we study the effect this choice has on numerical reasoning through the use of arithmetic tasks. We consider left-to-right and right-to-left tokenization for GPT-3.5 and -4, finding that right-to-left tokenization (enforced by comma separating numbers at inference time) leads to largely improved performance. Furthermore, we find that model errors when using standard left-to-right tokenization follow stereotyped error patterns, suggesting that model computations are systematic rather than approximate. We show that the model is able to convert between tokenizations easily, thus allowing chain-of-thought-inspired approaches to recover performance on left-to-right tokenized inputs. We also find the gap between tokenization directions decreases when models are scaled, possibly indicating that larger models are better able to override this tokenization-dependent inductive bias. In summary, our work performs the first study of how number tokenization choices lead to differences in model performance on arithmetic tasks, accompanied by a thorough analysis of error patterns. We hope this work inspires practitioners to more carefully ablate number tokenization-related choices when working towards general models of numerical reasoning.
\end{abstract}

\begin{figure}[h]
    \centering
    \includegraphics[width=\columnwidth]{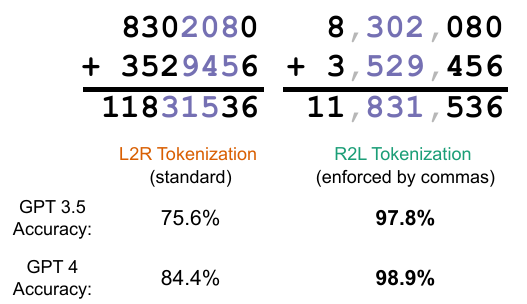}
    \vspace{-2em}
    \caption{Illustrating the dependence of frontier model arithmetic performance on tokenization. We show how using commas can enforce right-to-left (R2L) tokenization for the same addition problem. R2L tokenization leads to improved model performance on both GPT-3.5 and GPT-4 (March 2023 models), which we show is due to tokenization alignment between addends and answer through various controls and error analyses.}
    \label{fig:main_result}
    \vspace{-1.5em}
\end{figure}

\section{Introduction}

Large language models (LLMs) are often lauded as demonstrating the benefits of end-to-end learning over inductive biases. However, an often overlooked part of the pipeline, preventing it from being end-to-end, is tokenization: the segmenting of an input sequence of bytes into discrete tokens. Tokenization consists of two halves: training, in which a vocabulary of tokens and statistics are learned over a given corpus, and segmenting, where a function uses the trained vocabulary and statistics to map sequences of bytes to tokens. Each tokenization scheme may impart different inductive biases on the model due to the way in which bytes of input sequences are grouped -- in this work, we study these tokenization-dependent effects on numerical reasoning in state-of-the art models (GPT-3.5, GPT-4) by considering the tokenization of numbers in arithmetic problems.

\begin{figure*}[ht]
    \centering
    \includegraphics[width=\textwidth]{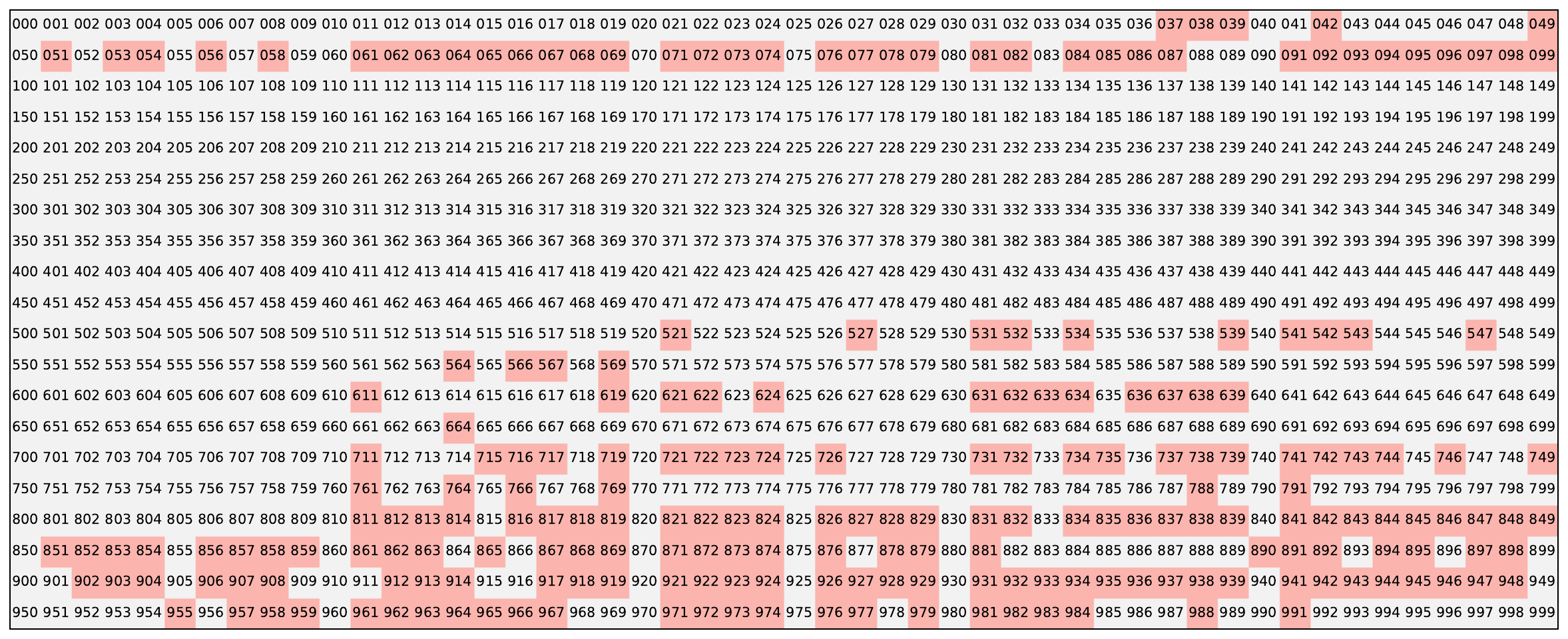}
    \vspace{-2em}
    \caption{All 3-digit strings, colored red when the string does not have a corresponding single token in \texttt{p50k\_base}, the BPE tokenizer for GPT-3. Though there's some patterns (e.g., nearly all multiples of 10 are present), overall there's no clear structure. The missing tokens are an artifact of the specific process BPE tokenizers use to establish vocabularies.}
    \label{fig:p50k_tokens}
    \vspace{-1em}
\end{figure*}

\begin{figure}[b!]
    \vspace{-2em}
    \centering
    \includegraphics[width=\columnwidth]{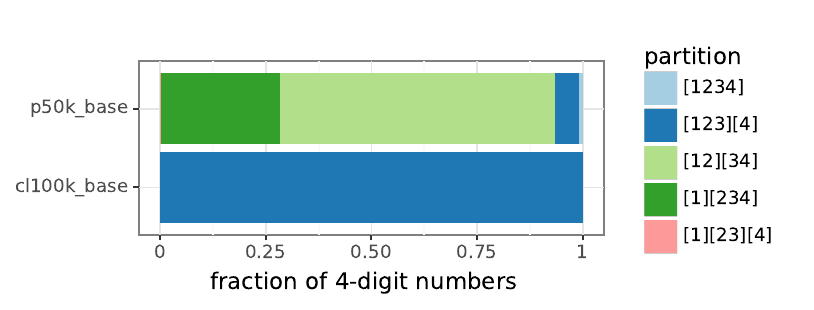}
     \vspace{-2em}
    \caption{Comparison of how \texttt{p50k\_base}, the tokenizer for GPT-3, and \texttt{cl100k\_base}, the tokenizer for GPT-3.5 and GPT-4, segments 4 digit strings into tokens. \texttt{cl100k\_base} standardized number tokenization to chunks of 3 digits, left-to-right, resulting in all N-digit numbers being segmented the same way.}
    \label{fig:4digit_tokenization_split}
\end{figure}

Though many techniques have been proposed for tokenization, the prevailing methods in today's frontier models are variants of Byte Pair Encoding (BPE) \citep{gage1994bpe, sennrich2016bpe}. BPE is a statistical approach to tokenization that is learnt from a dataset of, in the case of LLMs, text. Intuitively, BPE compresses the dataset by iteratively creating tokens for the most commonly occurring subsequences. Specifically, BPE begins with a token vocabulary consisting of each character in the text (e.g. letters, numbers, punctuation).\footnote{More precisely, single bytes are used to handle multilinguality, but this level of description suffices for our needs.} Using this vocabulary, bigram statistics (i.e. frequencies of pairs) are calculated, and the most common bigram is added to the vocabulary and merged in the dataset. This process is repeated up until a prespecified, maximum vocabulary size is reached. After the tokenizer is learned, tokenization of new text proceeds by iteratively merging characters/tokens in the same order as learned on the training dataset.

Naively applying BPE on internet-scale corpora leads to very idiosyncratic number tokenization \cite{teknium2023bpe_weird}. In the training phase, which numbers receive dedicated tokens is very adhoc -- for example, $710$ may get a dedicated token, while $711$ will not (Figure~\ref{fig:p50k_tokens}). In the segmenting phase, these adhoc tokens will lead to different partitionings of numbers of the same length. In Figure~\ref{fig:4digit_tokenization_split}, we illustrate the different partitionings of all 4 digit numbers when using the \texttt{p50k\_base} tokenizer that was used to train GPT-3 \cite{brown2020gpt3}. To control for effects on downstream performance from such idiosyncratic tokenization, prior work \citep{nye2021scratchpad, zhou2022algo_prompting} has used formatting, such as spaces or commas, to separate individual digits, ensuring each digit maps to a single token.

Newer models, and their corresponding tokenizers, indicate that LLM practitioners across different labs have also tried to control for idiosyncratic tokenization (Table~\ref{tab:tokenization_strategies}).\footnote{A noteable exception are Anthropic's Claude models, which still use pure BPE number tokens (see Appendix~\ref{appx:llm_tokenization_schemes}).} The PaLM \cite{chowdhery2022palm}, LLaMa \cite{touvron2023llama1} and Mistral\footnote{Verified by inspecting the tokens from huggingface \url{https://huggingface.co/docs/transformers/main/en/model_doc/mistral}.} \cite{jiang2023mistral} models switch to single-digit tokenization, similar to that enforced by \citet{nye2021scratchpad}. Interestingly, GPT-3.5 and GPT-4's tokenizer, \texttt{cl100k\_base}, introduces tokens for all 1-, 2-, and 3-digit strings.\footnote{This tokenization is enforced by the cryptic \texttt{pat\_str} parameter in their tokenization library, \url{https://github.com/openai/tiktoken/blob/main/tiktoken_ext/openai_public.py} Line 76.} Tokenization of numbers by these GPT models defaults to breaking a long number into 3-digit chunks, left-to-right, which we hypothesize (and later show) may create issues for numerical reasoning.

These varied approaches to tokenization by today's frontier LLMs indicate a lack of convergence in the field on best practices and call for a deeper analysis of (positive or negative) inductive biases imparted by various tokenization schemes. In this work, we provide the first systematic comparison of model performance on the same numerical reasoning tasks with varied tokenization. Specifically, we consider the latest GPT models on few-shot arithmetic tasks. We vary the tokenization direction to be the default left-to-right (L2R) or right-to-left (R2L). We find that model accuracy is up to 20\% higher when using R2L tokenization (Figure~\ref{fig:main_result}, Section~\ref{sec:R2L_overall}). We then provide a thorough analysis of error patterns across these two tokenizations (Section~\ref{sec:error_patterns_overall}). We find that the difference in performance between R2L and L2R tokenization in GPT-3.5 can largely be explained by an extremely stereotyped and surprising error pattern (Section~\ref{sec:error_digit4}), perhaps indicating the presence of some systematic, but flawed, reasoning. Next, we show that chain-of-thought-inspired approaches, where a model is asked to repeat an input in R2L tokenization, recover the accuracy otherwise lost due to L2R tokenization (Section~\ref{sec:repeat_experiments}). Finally, we conclude by studying how these effects may change with model version, finding that larger models are better able to override the tokenization-induced effects but, as of yet, unable to eliminate them (Section~\ref{sec:later_models}). Overall, we view these results as compelling evidence towards significant tokenization-dependent inductive biases in large language models, and hope they lead model practitioners to conduct careful pre-training ablations with varying tokenization schemes, especially for numerical reasoning.

\begin{table}
\caption{Popular LLMs and their number tokenization strategies. BPE = byte pair encoding. L2R = left-to-right.}
\center{
\begin{tabular}{l l}
\toprule
\textbf{Model}   & \textbf{Strategy}      \\ \midrule
GPT-3 (2020)        & pure BPE               \\ %\midrule
GPT-3.5 (2022)      & L2R chunks of 3 digits \\ %\midrule
GPT-4 (2023)        & L2R chunks of 3 digits \\ %\midrule
Claude v2.1 (2023)  & pure BPE               \\ %\midrule
Gopher (2021)       & pure BPE               \\ %\midrule
Chinchilla (2022)   & pure BPE               \\ %\midrule
PaLM (2022)         & single digit           \\ %\midrule
\hline
GPT-J (2021)        & pure BPE               \\
Llama 1 \& 2 (2023) & single digit           \\ %\midrule
Mistral (2023)      & single digit           \\ 
OLMo (2024)         & pure BPE               \\ \bottomrule
\end{tabular}
}
\vspace{-2em}
\label{tab:tokenization_strategies}
\end{table}

\section{Methods}
\label{sec:methods_overall}

\subsection{Experiment setup}
\label{sec:experiment_setup}

We evaluate GPT models through the Chat Completions endpoint on the OpenAI API\footnote{\url{https://platform.openai.com/}} on few-shot addition problems. We control for addend digit length, ranging from 7 to 9 digits (chosen since this way each addend is 3 tokens long). For most experiments, we use 90 random problems, with 10 problems for each addend digit length pair (e.g., 10 problems where the addends are both 7 digits long, 10 problems where the first addend is 7 digits and the second is 8, etc.). For shots, we consider 1-, 2-, 4-, and 8-shots. Shots are sampled randomly for each ``query'' problem, and are provided to the model as a multi-turn dialogue. We control shots to have the same form (digit lengths, tokenization direction, etc.) as the query problem. We use the default system prompt ``You are a helpful assistant.'' for maximum reproducibility.\footnote{We did experiment with other system prompts and found minimal differences (Appendix~\ref{appx:mathgpt_prompt}).} Python code for some example 2-shot queries to the model are presented in Appendix~\ref{appx:prompts} for maximum clarity. We use greedy decoding (temperature=0) in all experiments. Accuracy was computed by extracting numbers from model responses.

Most experiments in the paper are run using the \texttt{gpt-3.5-turbo-0301} model checkpoint, though in Section~\ref{sec:later_models} we look into how results extend to newer versions of the same model (\texttt{gpt-3.5-turbo-0613}) and to the, presumably larger, GPT-4 models (\texttt{gpt-4-0314, gpt-4-0613}). All code and full results tables can be 
found at \url{https://github.com/aadityasingh/TokenizationCounts}.

\subsection{Varying L2R vs. R2L tokenization}
\label{sec:r2l_methods}

The ChatCompletion API only allows for input text, not input tokens, so it's tricky to conduct tokenization-varying experiments. To force the model to use R2L tokenization for numbers, we add commas every 3-digits from the right (see Figure~\ref{fig:main_result}). Since the tokenizer doesn't contain any tokens with numbers and commas, the commas get tokenized separately, effectively enforcing a different segmentation of digits. We use this setting to illustrate our main results, and conduct various controls to ensure that our observed effect is due to tokenization as opposed to other confounds.

\section{Right-to-left tokenization improves model performance}
\label{sec:R2L_overall}

\subsection{Main results}
\label{sec:comma_R2L}

\begin{figure}[ht]
    \centering
    \vspace{-0.8em}
    \includegraphics[width=\columnwidth]{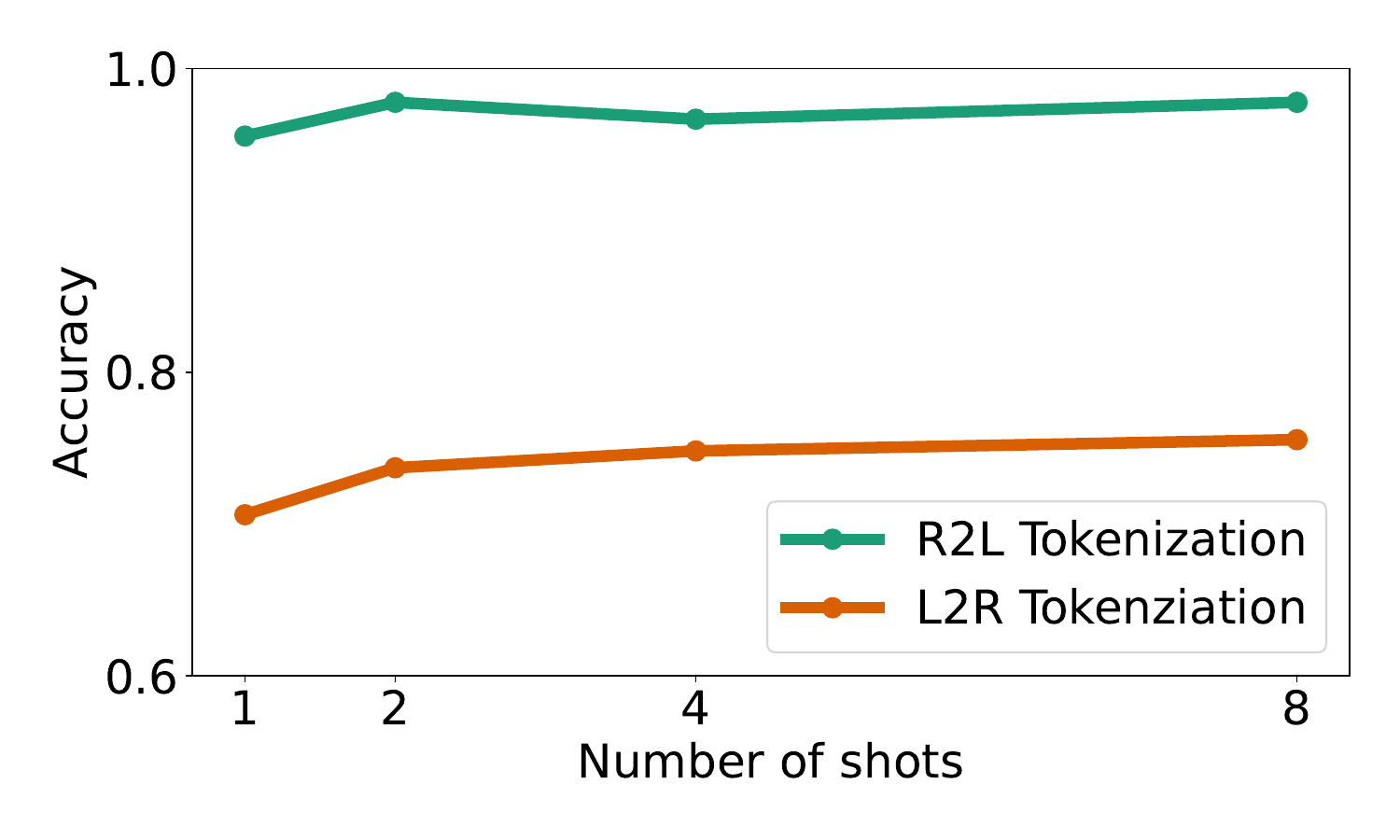}
    \vspace{-2em}
    \caption{Effect of R2L vs L2R tokenization with increasing shots.}
    \label{fig:token_dir_shots}
    \vspace{-1em}
\end{figure}

When using commas to separate digits and enforce R2L tokenization, we observed greatly improved average performance (8-shot result in Figure~\ref{fig:main_result}). We found that increasing the number of shots (Figure~\ref{fig:token_dir_shots}) led to a larger increase for the L2R tokenization (from 68.5\% 1-shot to 75.6\% 8-shot) than for the R2L tokenization (from 95.6\%  1-shot to 97.8\% 8-shot) indicating that in-context learning may slightly mitigate the (harmful) bias of L2R tokenization. Given this finding and the plateau-ing in performance with increasing shots, we report only 8-shot results for the remainder of the work as this makes L2R tokenization the most competitive.

\subsection{Controlling for comma-based semantic priors}
\label{sec:control_delim}

Though this result is already compelling, we realize that commas are often used to separate digits in the manner depicted in Figure~\ref{fig:main_result}, so the observed effect may be confounded by prevalence in training data \cite{mccoy2023embers}. One might argue that comma separation is actually bringing the input closer to the training distribution of the model, so it's not a surprise that models perform better. To control for this and focus in on tokenization, we consider alternate, single-token separators: \texttt{' ', '.', '\$', '\#'} (note we'll refer to \texttt{' '} as \texttt{<space>} for clarity). For example, the number \texttt{8302080} would be written as \texttt{8\#302\#080} when input to the model.

\begin{figure}[ht]
    \centering
    \vspace{-1em}
    \includegraphics[width=\columnwidth]{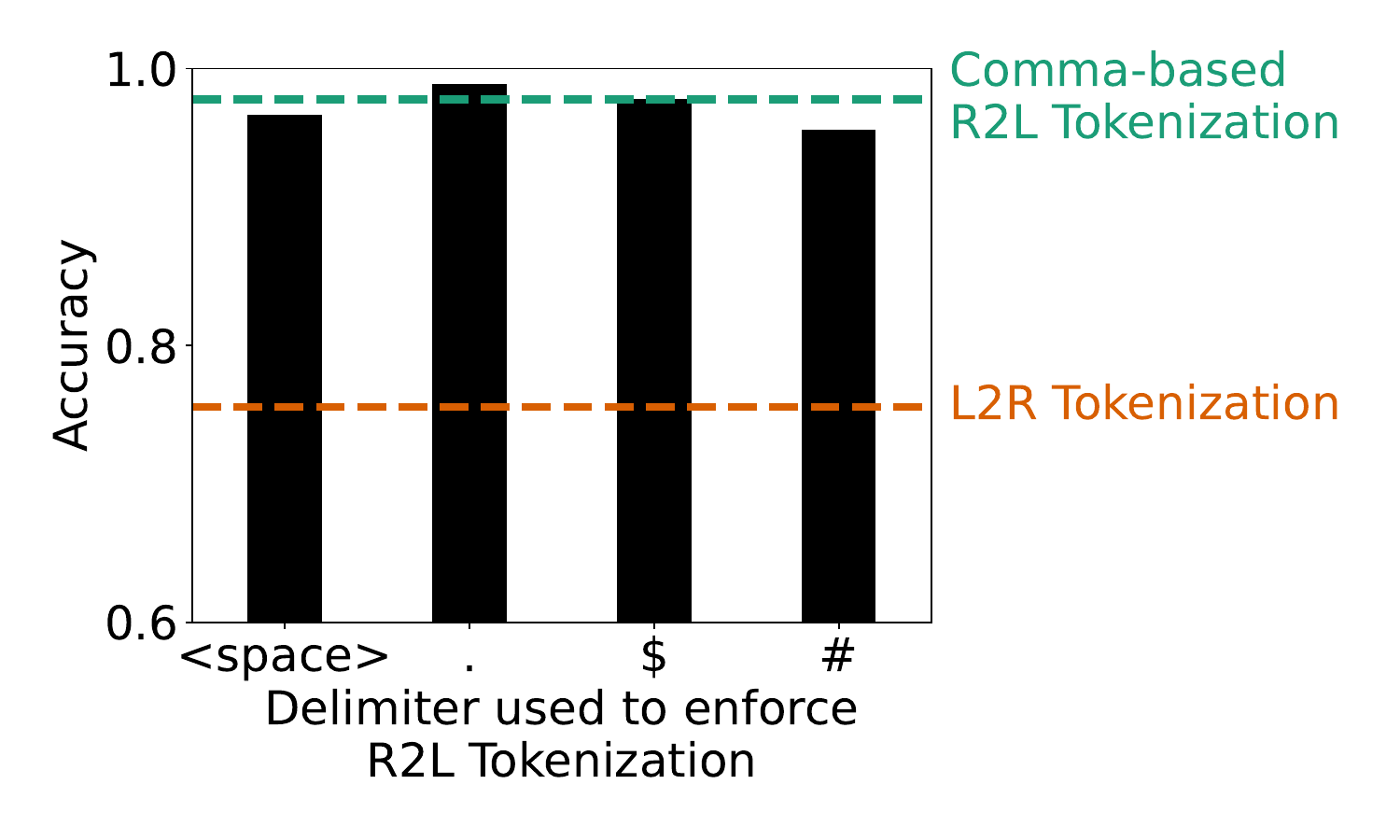}
    \vspace{-3em}
    \caption{8-shot accuracy when using different delimiters for R2L tokenization. Dotted lines show results from Figure~\ref{fig:main_result} for comparison. Overall, we see choice of delimiter matters less than direction of tokenization.}
    \vspace{-1em}
    \label{fig:control_delim}
\end{figure}

Results are shown in Figure~\ref{fig:control_delim}. We find that the model is largely agnostic to the separator used, indicating that tokenization is likely the dominant effect, rather than the specific choice of using commas.

\subsection{Controlling for ``thinking tokens''}
\label{sec:thinking_tokens}

Another confound with the above experiment may be that adding commas both increases the number of tokens input to as well as generated by the model. Thus, to generate the same answer, the model has access to more computation steps (i.e., FLOPs). There is a worry that models may use these repetitive \textit{thinking tokens} to perform additional useful computations \citep{lanham2023faithfulness}. In practice, this seems not to happen without further training \citep{goyal2024pause}, but we conducted experiments to verify this in our setting.

To control for thinking tokens, we consider two types of controls. In the first, we use separators to enforce L2R tokenization -- this enforces an exact match in prompt token counts. Second, we consider adding 1 or 2 spaces before and after the \texttt{+} and \texttt{=} sign to increase the number of tokens\footnote{Specifically, the token count in the 8-shot prompt increases from 195 to 213 to 240 when going from 0 to 1 to 2 spaces. For reference, the R2L token count is 247.} in the L2R case (where no separator is used). Both of these have the benefit of adding extremely ``predictable'' tokens (when using 8-shots), allowing the model to possibly use the extra computation steps for ``thinking''.

\begin{figure}
    \centering
    \vspace{-0.8em}
    \includegraphics[width=\columnwidth]{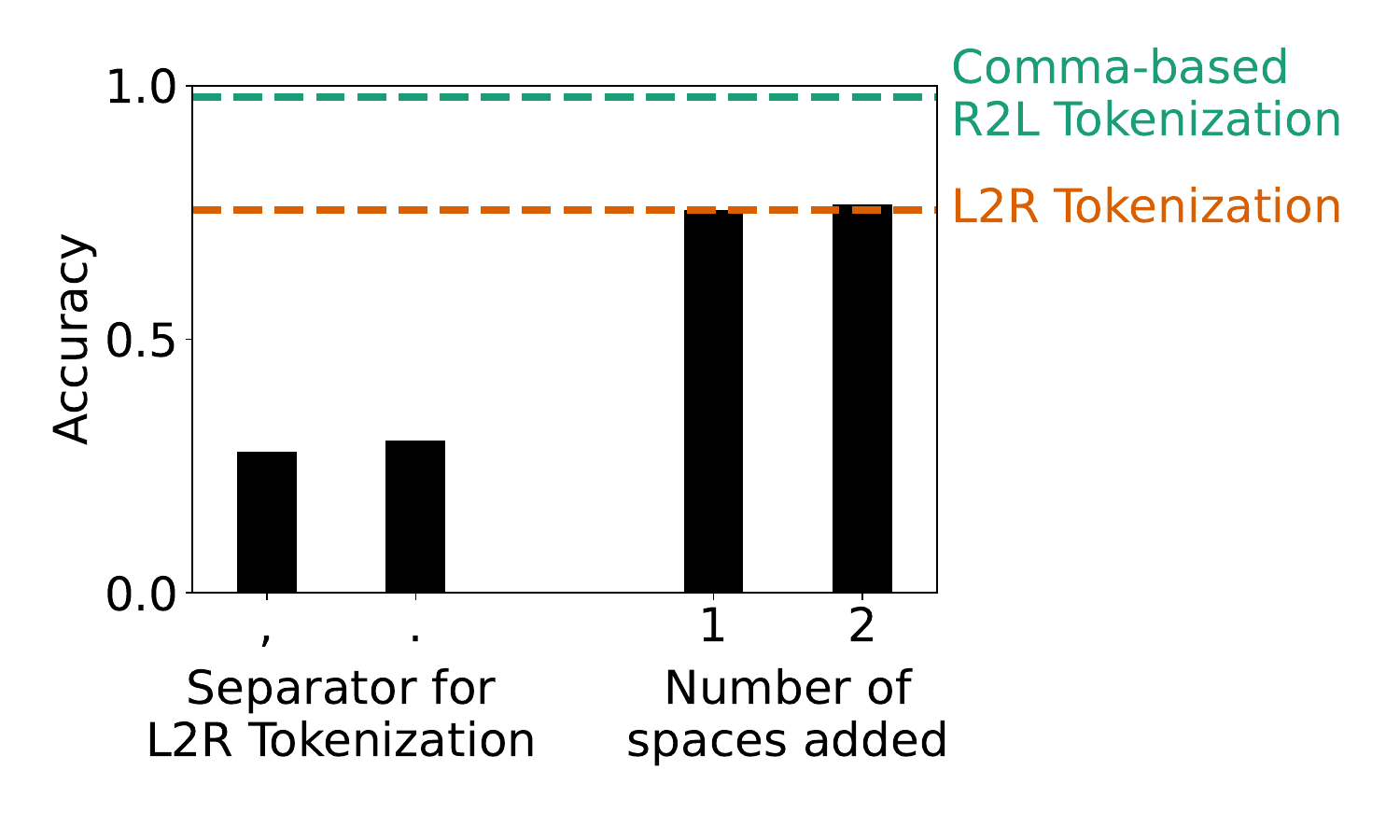}
    \vspace{-3em}
    \caption{8-shot accuracy for various ``thinking token'' controls. Dotted lines show results from Figure~\ref{fig:main_result} for comparison. We also experimented with other delimiters for L2R tokenization (all those from Figure~\ref{fig:control_delim}), but found similarly poor results. Overall, ``thinking tokens'' do not recover the performance boost from using comma-enforced R2L tokenization.}
    \vspace{-1em}
    \label{fig:control_thinking_token}
\end{figure}

In Figure~\ref{fig:control_thinking_token}, we find that neither of these controls, when applied to L2R tokenized sequences, recovers the performance of R2L tokenization. In fact, we found that using separators with the L2R tokenization often hurt performance, likely because this is an uncommon representation---upon qualitative inspection of a few examples, we found the model sometimes ``auto-corrects'' the inputs by hallucinating trailing zeros. We believe these experiments effectively rule out the ``thinking token'' confound.

\section{Error analysis reveals stereotyped patterns}
\label{sec:error_patterns_overall}

Given the robust effect observed in Section~\ref{sec:R2L_overall}, we were curious to see if there were any patterns in the errors. Below, we summarize our key findings.

\subsection{L2R tokenization is significantly worse when answer is longer than addends}
\label{sec:error_length}

As noted in Section~\ref{sec:experiment_setup}, we balanced our dataset of problems based on input digit length. Upon inspection of problems the model got incorrect when using L2R tokenization, we noted that errors seemed more likely when the answer was longer than the addends (e.g., a problem where 7 digit number + 7 digit number = 8 digit number, of the form depicted in Figure~\ref{fig:main_result}). To test this hypothesis, we conducted a new experiment where we controlled for addend lengths \textit{and} answer lengths. Specifically, we generated 100 random problems for each possible triplet of digit lengths where addends and answer have a length of 7 to 9 digits (full list in Appendix~\ref{appx:length_control}). The remainder of our experiments in this section will use this expanded set of problems to show the robustness of the found error patterns.

\begin{figure}
    \centering
    \includegraphics[width=\columnwidth]{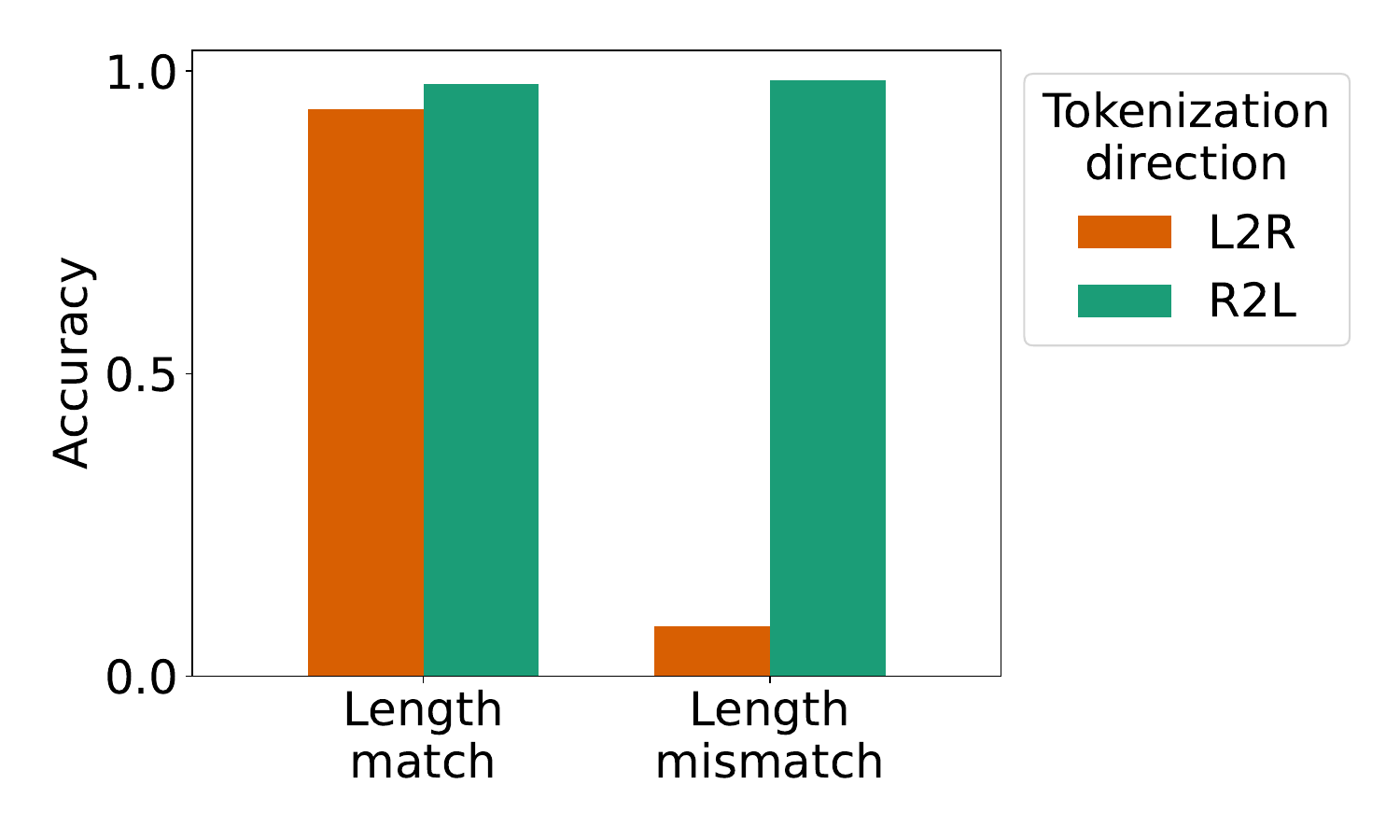}
    \vspace{-3em}
    \caption{When the answer is the same length in digits as an addend (length match), both tokenization schemes perform similarly (left). When the answer is a different length in digits than either addend (length mismatch), L2R tokenization destroys model performance, dropping to 8.25\% (right).}
    \vspace{-0.5em}
    \label{fig:answer_length}
\end{figure}

\begin{figure}
    \centering
    % \vspace{-0.5em}
    \includegraphics[width=\columnwidth]{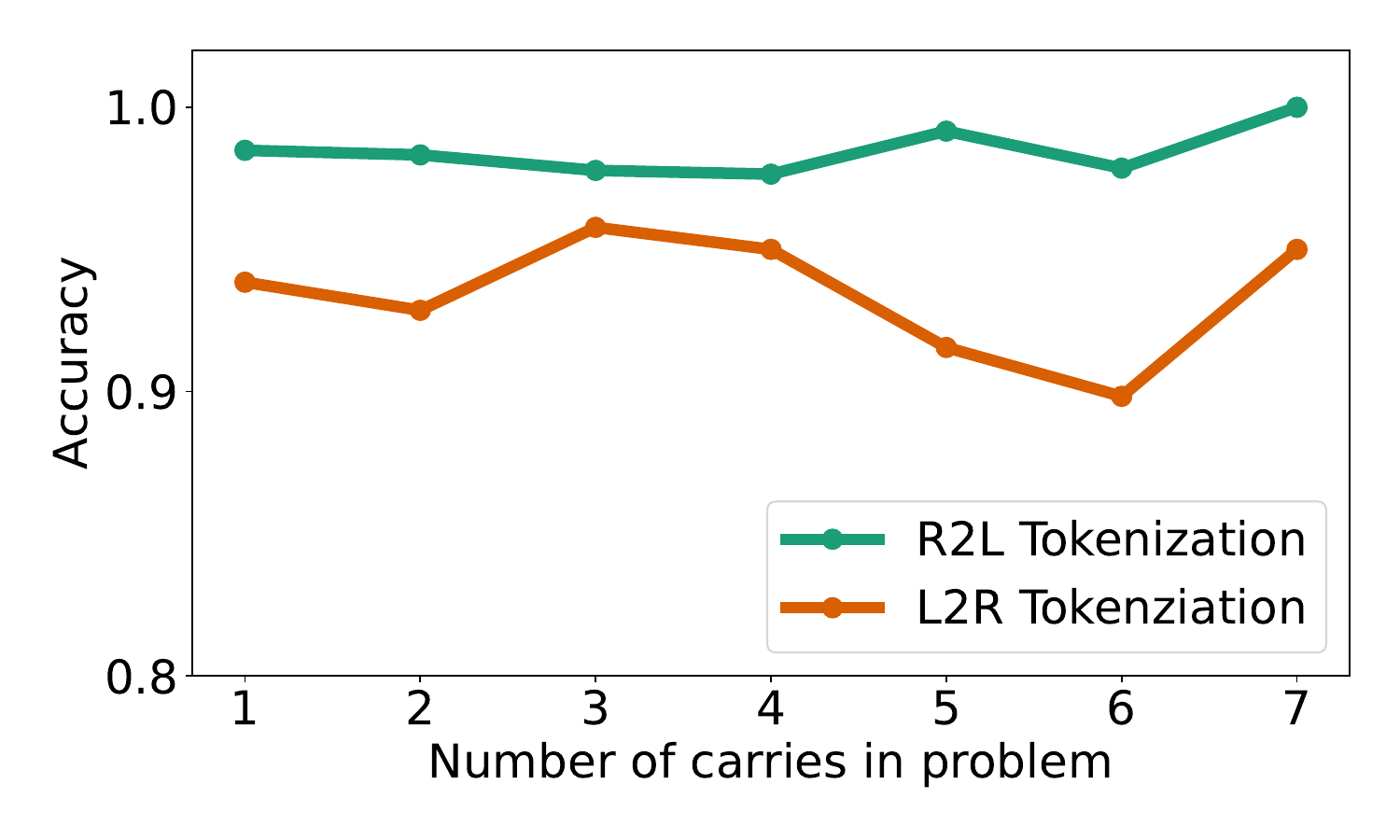}
    \vspace{-3em}
    \caption{Accuracy as a function of number of carries. For L2R tokenization, we exclude problems where the answer length does not match at least one addend, as the model misses most of those (92\%) as shown in Section~\ref{sec:error_length}. If number of carries (a human notion of difficulty) was correlated to model performance, we would expect a negative slope. The lack of any trend suggests model performance is largely independent of number of carries.}
    \vspace{-1.5em}
    \label{fig:carries}
\end{figure}

We reproduced our main phenomenon (Section~\ref{sec:comma_R2L}), and further affirmed our intuitions about error patterns. As shown in Figure~\ref{fig:answer_length}, we find that L2R tokenization has similar performance to R2L tokenization when the answer's length in digits is the same as one of the inputs (which we refer to as the ``length match'' condition). When the answer is longer than the inputs (due to a final carry), L2R tokenization is significantly worse, with accuracy dropping down to 8.25\% -- we refer to this as the ``length mismatch'' condition. We suspect that this strong effect may be due to the misalignment between input and output tokenizations (as illustrated in Figure~\ref{fig:main_result}) rather than some carry-related notion of problem difficulty, which we explore in the next few subsections.

\subsection{Errors do not seem correlated to number of carries}

A natural hypothesis given the above result may be that errors might just be correlated to some notion of difficulty, such as carries. In Figure~\ref{fig:carries}, we find that this is generally not the case. Specifically, we consider the accuracy on subsets of problems based on how many carries are needed to solve them.\footnote{Since we didn't explicitly control for carries when generating problems, the number of problems with a given number of carries varies. We only considered cases where we had at least 50 problems.} The lack of a clear positive or negative trend indicates that model performance is not strongly affected by the number of carries.

\subsection{Length mismatch problems yield stereotyped ``digit 4'' error pattern}
\label{sec:error_digit4}

\begin{figure}[ht]
    \centering
    \vspace{-0.5em}
    \begin{subfigure}[t]{0.95\columnwidth}
        \caption{Location of errors}
        \includegraphics[width=0.95\columnwidth]{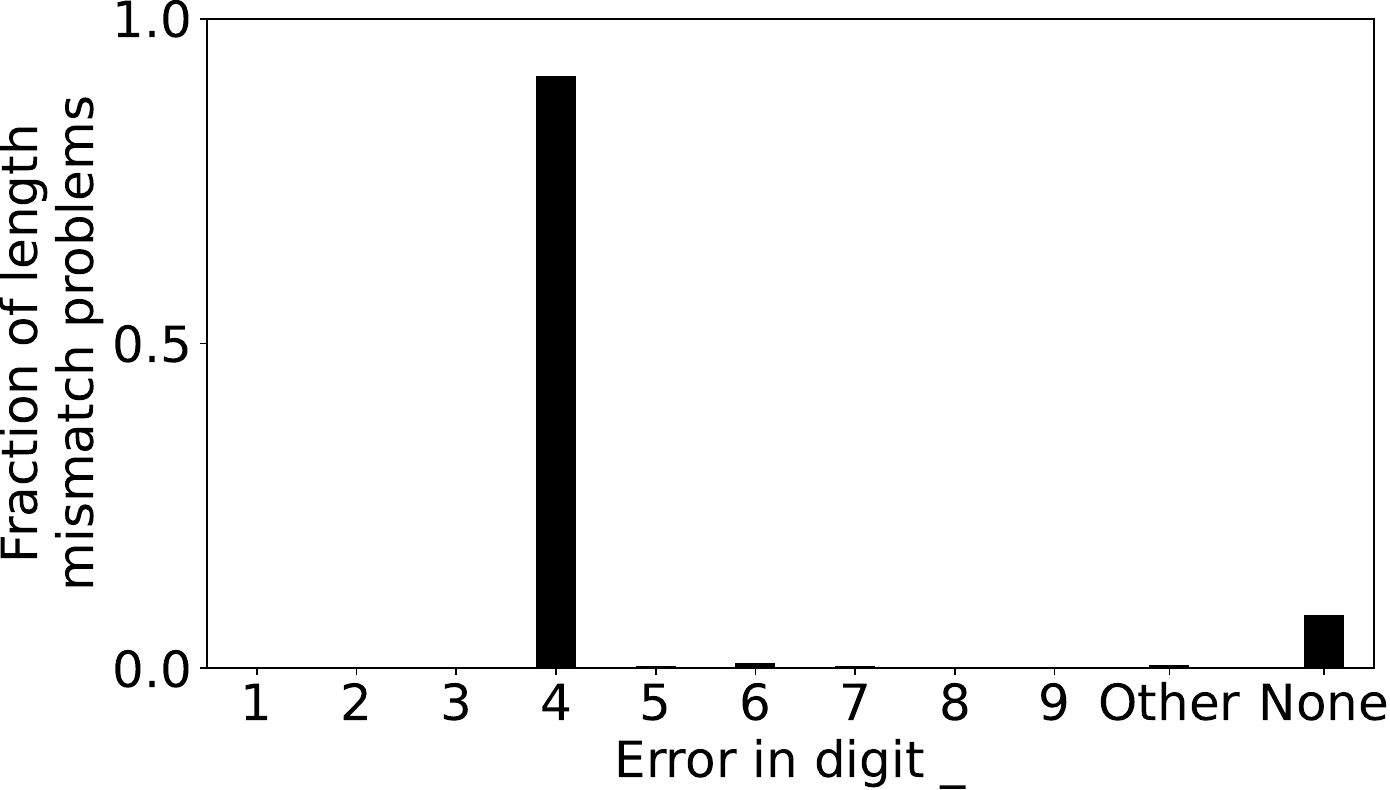}
    \end{subfigure}
    \begin{subfigure}[t]{0.96\columnwidth}
        \vspace{-0.3em}
        \caption{Magnitude of errors}
        \includegraphics[width=0.95\columnwidth]{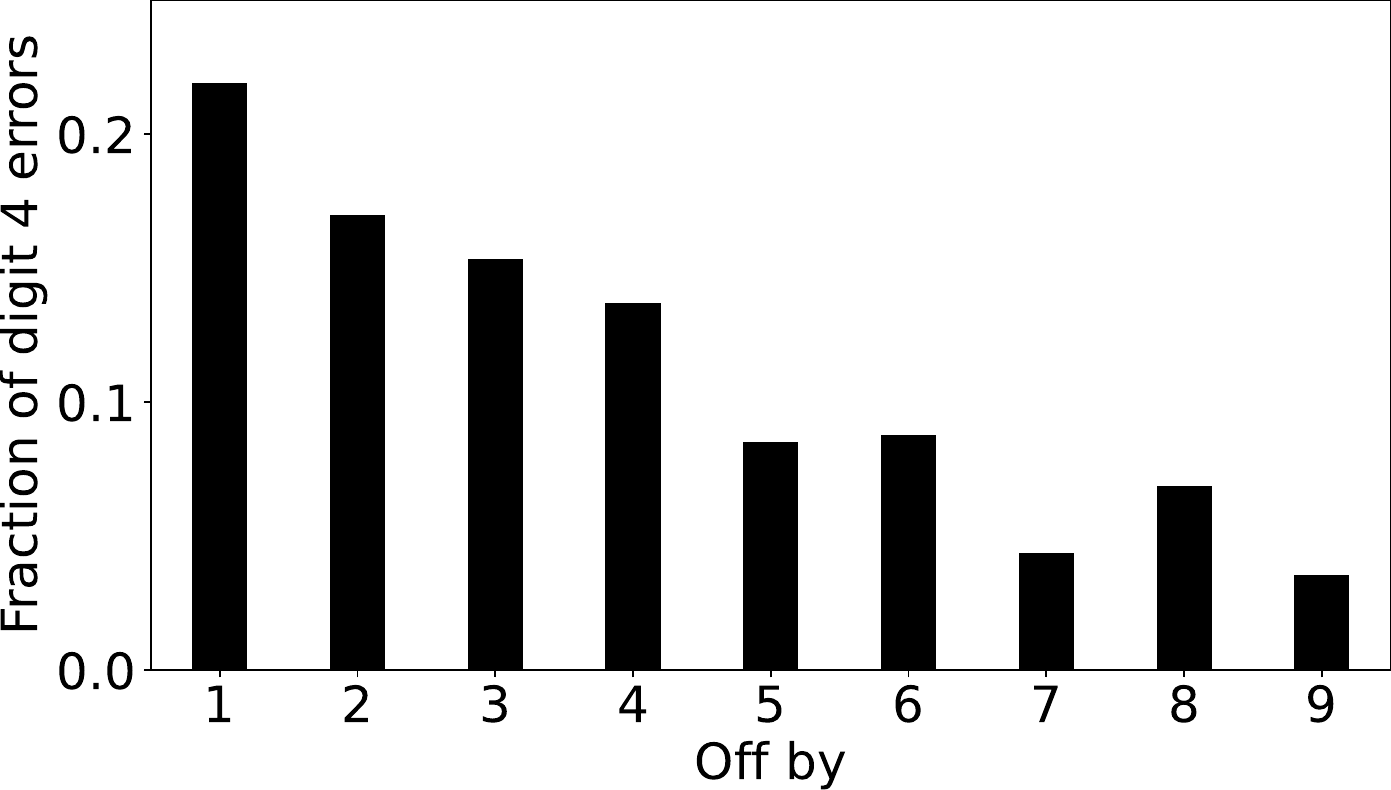}
    \end{subfigure}
    \vspace{-0.5em}
    \caption{\textbf{a)} Error patterns for L2R tokenization over problems where the answer digit length is different than the addend lengths. ``None'' indicates the problems in this case that the model got correct (8.25\%). ``Other'' indicates problems where the model doesn't provide a valid answer or provides an answer of the wrong length (0.5\%). Of the remaining 91.25\% which the model gets incorrect, it shockingly \textit{always} gets digit 4 wrong. In addition, it sometimes gets other digits (5, 6 or 7) wrong. \textbf{b)} For the errors in digit 4, we show the magnitude of the mistake. For example, if the correct value of digit 4 is \texttt{2} and the model response has digit 4 equal to \texttt{5}, it would be off by \texttt{3}. We see a slight preference to off-by-1 errors, but  error magnitudes are fairly evenly distributed.}
    \vspace{-1.5em}
    \label{fig:error_digit4}
\end{figure}

If not carries, what could be causing the surprising error pattern in Figure~\ref{fig:answer_length}? In Figure~\ref{fig:error_digit4}a, we find that the errors when using L2R tokenization are extremely stereotyped and not at all intuitive. Specifically, in the length mismatch condition, the model \textit{always} gets the fourth digit wrong. Furthermore, the model \textit{always} gets the first 3 digits correct (corresponding to the first output token). In terms of how far off the model is on digit 4, Figure~\ref{fig:error_digit4}b shows that there's a slight preference to off-by-one errors, but overall the specific substitution appears quite haphazard. 

We found this result extremely surprising. In cognitive science, such stereotyped error patterns are often used as evidence of underlying systematic processing. While the mechanism for addition in LLMs remains unclear, we find this striking, tokenization-dependent error pattern\footnote{This error pattern is not present when using R2L tokenization.} as highly suggestive of some underlying algorithm (in contrast to suggestions that LLMs may be performing arithmetic using some ``fuzzy'' matching to similar problems in training). We provide further evidence of stereotyped error patterns by analyzing model log probabilities in Appendix~\ref{appx:logprobs}.

\subsection{Off-by-one errors at token boundaries account for nearly all remaining errors}
\label{sec:errors_rest}

After accounting for the main source of error, we analyzed the remaining errors across both tokenization methods: 25 out of the 1300 problems for R2L tokenization, and 56 out of the 900 problems in the length match condition for L2R tokenization.

For R2L tokenization, of the 25 problems missed, 24 are due to off-by-one (either above or below) errors. For L2R tokenization, of the 56 problems missed, 53 are due to off-by-one (either above or below) errors. For nearly all these off-by-one errors,\footnote{Specifically, all 24 off-by-one errors in the R2L case, and 51 of the 53 off-by-one errors in the L2R case.} regardless of tokenization direction, we find that the error itself occurs in \textit{the last digit of an output token}. This result suggests that off-by-one errors are more likely across token boundaries as opposed to in the middle of a 3-digit token. This hypothesis, with preliminary evidence, connects to works on length generalization \cite{anil2022exploring} -- using 3-digit tokens may make length generalization easier as models only need to cross token boundaries every third digit (as opposed to every digit).

\section{Models are able to convert from L2R to R2L tokenization, improving performance}
\label{sec:repeat_experiments}

\subsection{Main results}
\label{sec:repeat_experiments_main_results}

\begin{figure}[ht]
    \centering
    \vspace{-1em}
    \includegraphics[width=\columnwidth]{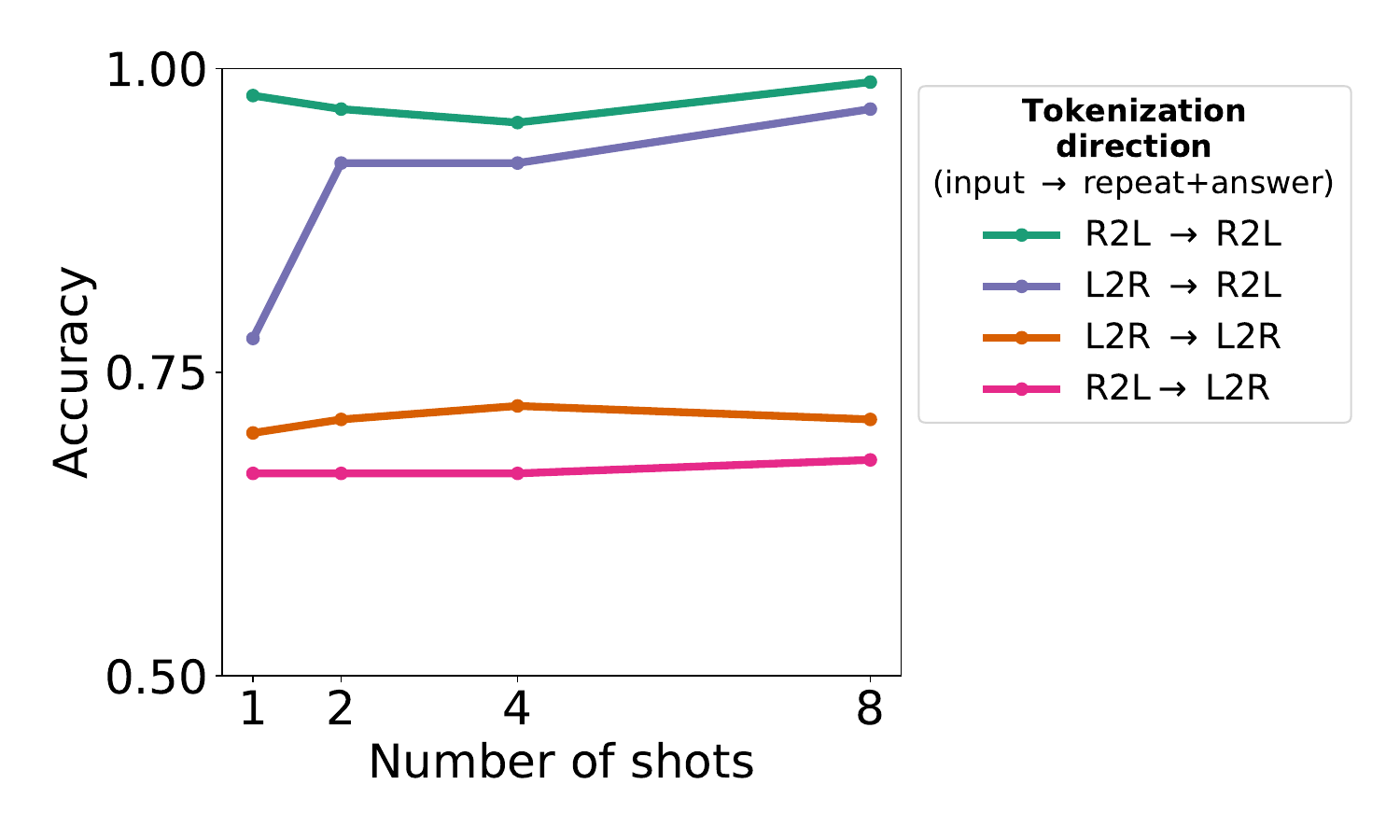}
    \vspace{-3em}
    \caption{Few-shot accuracy when models receive a problem with one tokenization direction, then repeat and answer it in another.}
    \vspace{-1em}
    \label{fig:repeat_style}
\end{figure}

With the above results showing that number tokenization can strongly affect numerical reasoning, we ask if models can be prompted to take problems in a less preferred tokenization (L2R) and convert them to a more preferred tokenization (R2L) to improve performance. Inspired by chain-of-thought approaches \cite{nye2021scratchpad, kojima2022cot, wei2022cot}, we few-shot prompt models to take problems with one tokenization direction, and then repeat the problem and answer it using a different tokenization direction. In Figure~\ref{fig:repeat_style}, we find that models indeed perform nearly as well at addition when converting L2R tokenization to R2L themselves as to when they receive the problem in R2L tokenization in the first place. Performance increases with the number of shots when converting L2R to R2L since the model adheres more to the (helpful) suggested repetition style. These results indicate that models can convert between tokenizations to solve problems correctly, but do not do so implicitly in the forward pass.

\subsection{Controlling for output tokenization}
\label{sec:repeat_output_control}

\begin{figure}[ht]
    \centering
    \includegraphics[width=\columnwidth]{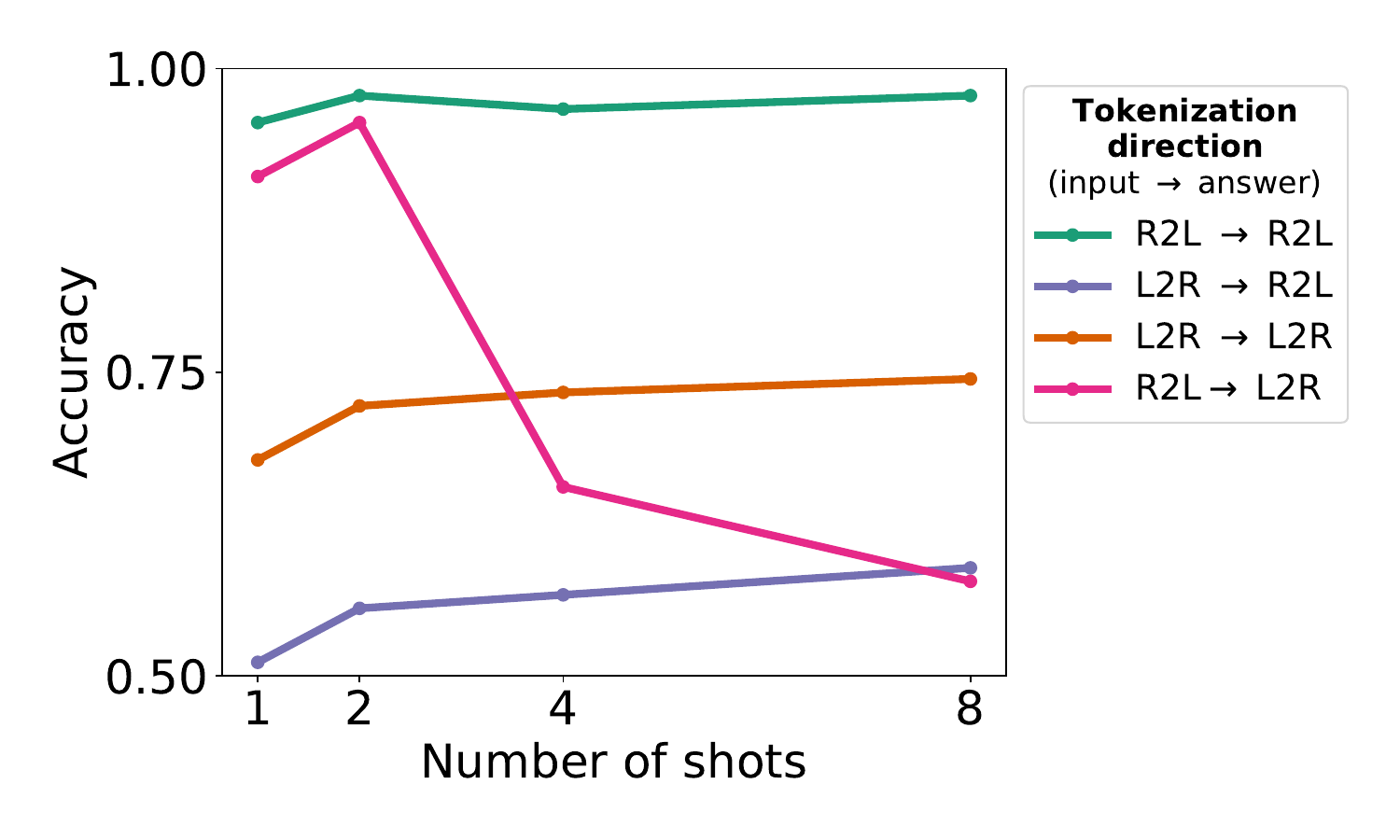}
    \vspace{-3em}
    \caption{Few-shot accuracy when models receive a problem in one tokenization format, and answer it in another. The distinction between this and Figure~\ref{fig:repeat_style} is that models \textit{do not} repeat the problem in this case. We note that when giving a model a problem in R2L tokenization and prompting it to answer in L2R tokenization, the model actually gets \textit{worse} with more shots, since for fewer shots, the model ends up ignoring the few-shot prompt and answers in its preferred R2L tokenization. Specifically, adherence to the prompted formatting for R2L$\rightarrow$L2R increases from just 13.3\% with 1 shot to 98.9\% with 8 shots.}
    \label{fig:answer_style}
\end{figure}

One confound with the above experiment may just be that the model improves when it's asked to generate \textit{answers} with R2L tokenization. To control for this, we conduct a similar experiment, but without few-shot prompting the model to repeat the problem: the few-shot prompt provides answers with a different tokenization direction than the input, incentivizing the model to answer with this tokenization direction (see Appendix~\ref{appx:prompts} for an example prompt). In Figure~\ref{fig:answer_style}, we see that just answering with R2L tokenization does not improve performance (purple curve) to the degree that repeating in R2L tokenization does (purple curve, Figure~\ref{fig:repeat_style}), when starting from L2R tokenization. This effect indicates that it is important for the model to also \textit{see} the problem in the preferred tokenization (by repeating it), rather than just answering in the preferred tokenization. 

\begin{figure}
    \centering
    \includegraphics[width=\columnwidth]{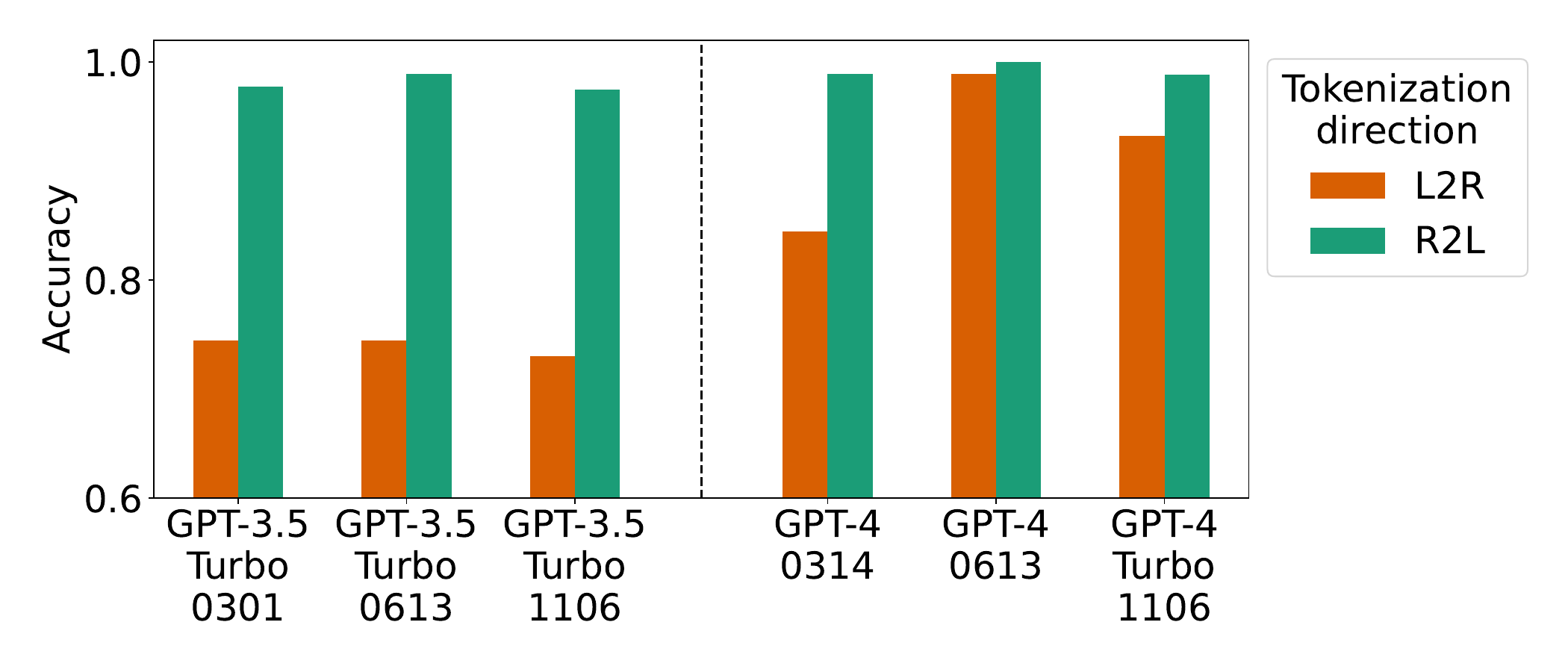}
    \vspace{-2em}
    \caption{8-shot performance of various OpenAI models on the same addition problems as Figure~\ref{fig:token_dir_shots}. The newer version sof GPT-3.5 appears to perform equally poorly. For GPT-4, we see a large tokenization dependent effect in the March model, which becomes weaker (but still present) in the June model. The GPT-4 turbo model shows a slight regression in overall performance with the tokenization-dependent effect becoming stronger again.}
    \vspace{-1em}
    \label{fig:main_effect_new_models}
\end{figure}

\section{Tokenization-dependent effects mostly extend to future models}
\label{sec:later_models}

Through the previous sections, we've demonstrated a strong tokenization-dependent effect. In this section, we address the question: does this effect extend to newer models?

\begin{figure}
    \centering
    \includegraphics[width=\columnwidth]{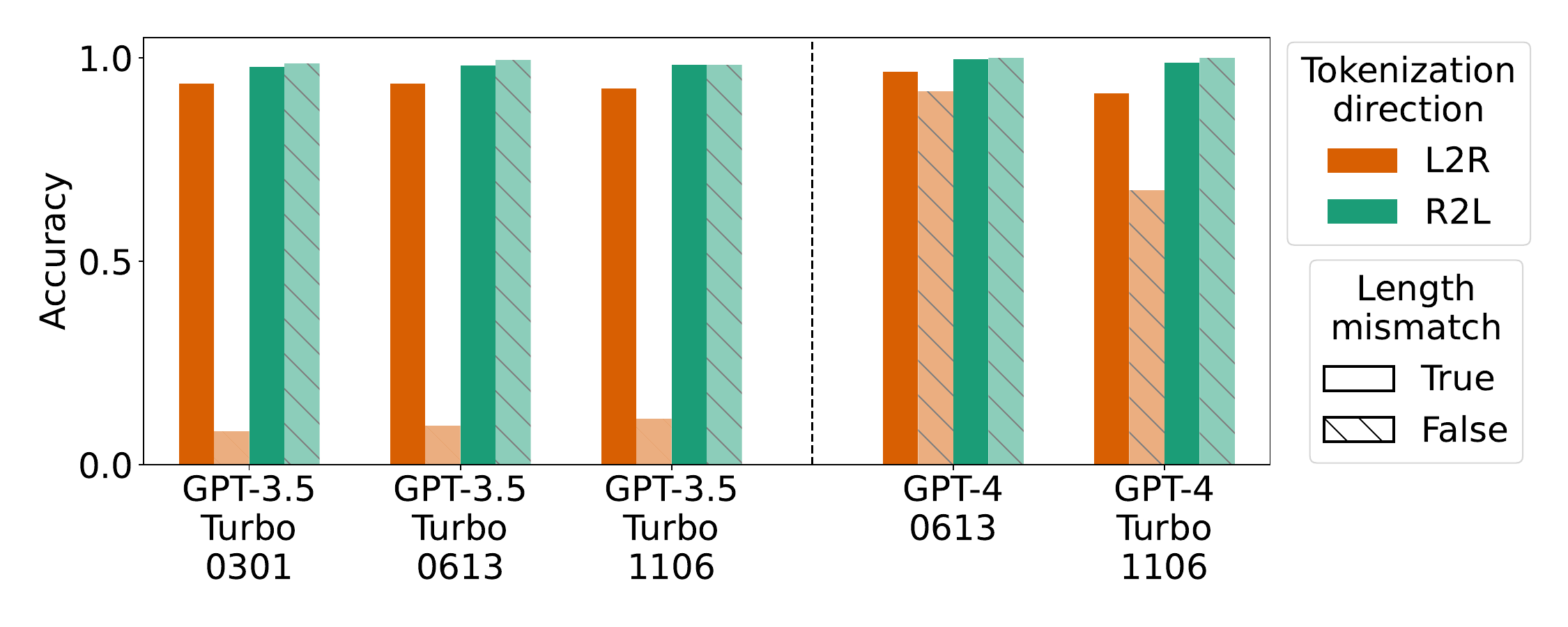}
    \vspace{-2.5em}
    \caption{8-shot performance of various OpenAI models on answer length controlled problems (see Section~\ref{sec:error_length}), separated by whether the answer length is the same as one of the addends. We see the effect from Section~\ref{sec:error_length} reproduces strongly in the newer version of GPT-3.5. The effect is still present in GPT-4,\footnote{We were unable to run experiments on \texttt{gpt-4-0314} for this analysis as the model has been removed from the API, despite a stated deprecation date of June 13, 2024} but not as strongly. Interestingly, the effect is stronger in the latest GPT-4 Turbo model as compared to GPT-4.}
    \vspace{-1.5em}
    \label{fig:error_new_models}
\end{figure}

As shown in Figure~\ref{fig:main_effect_new_models}, we find that generally, yes: tokenization-dependent effects persist. We consider five ``held-out'' OpenAI models, which allow us to consider how tokenization-dependent effects shift when models are updated (\texttt{gpt-3.5-turbo-0613}, \texttt{gpt-3.5-turbo-1106}) or scaled up (\texttt{gpt-4-0314}, \texttt{gpt-4-0613}) and then scaled back down (\texttt{gpt-4-1106-preview}, which is a ``turbo'' model\footnote{We assume this is a smaller, maybe distilled, version of GPT-4.}). Later versions of GPT-3.5 exhibit as strong an effect due to tokenization direction. The effect is mitigated slightly in GPT-4's March version, and mitigated strongly in GPT-4's most recent version.\footnote{We find it interesting that the March to June update to GPT-4 improved performance, but the corresponding update to GPT-3.5 did not -- without knowing what these updates entail, however, it's hard to draw conclusions as to why this may be the case.} Specifically, GPT-4 models appear to be better at performing arithmetic across the board (for both tokenization directions). Interestingly, in the most recent GPT-4 Turbo model, the effect of tokenization becomes stronger again. Furthermore, Figure~\ref{fig:error_new_models} shows that the digit length mismatch between answer and addends is again the main reason for the performance drop when using L2R tokenization, in both GPT-3.5 and GPT-4 models. We believe that the increased scale of training GPT-4 (likely in both parameter count and data seen) allows it to better override the tokenization-induced inductive bias that leads GPT-3.5 models to perform worse (analogous to scale helping mitigate tokenization-induced spelling difficulties \citep{liu2023character}). The resurgence of tokenization-dependent effects in the newest GPT-4 Turbo model (which is presumably smaller than GPT-4) supports this hypothesis.

\section{Related work}
\label{sec:related_work}

\textbf{Tokenization methods} The two leading tokenization methods are Unigram \cite{kudo2018unigram} and BPE \cite{sennrich2016bpe}. While older work in NLP show the benefits of Unigram over BPE \cite{bostrom2020bpe}, BPE remains the most commonly used tokenization method by modern LLM practitioners. Within BPE, different models often make different hard-coded choices, such as removing long tokens of consecutive whitespace \cite{touvron2023llama1} or enforcing single-digit tokenization of numbers \cite{chowdhery2022palm}. Our work demonstrates tokenization-dependent effects from one such choice, the use of 1-, 2-, and 3- digit tokens by OpenAI models \cite{openai2023gpt4}. One way around such issues could be tokenizer-free methods (e.g., MEGABYTE \cite{yu2023megabyte}, which uses patch-based schemes and doesn't assume fixed tokens), but we suspect these schemes will also carry their own inductive biases. \citet{golkar2023xval} also introduce a continuous number encoding scheme meant to circumvent tokenization artifacts, but their approach is limited to cases where model outputs are purely numerical, and not interleaved with text.

\textbf{Tokenization artifacts in LLMs} A growing set of results has emerged around various tokenization-related artifacts in LLMs. Similar to scratchpad prompting \cite{nye2021scratchpad}, \citet{wei2023secondletter} found that separating letters into individual tokens can help in sorting words by the second letter. Other work \cite{shin2020autoprompt} has focused on specific tokens that can negatively affect model performance. \citet{rumbelow2023solidgoldmagikarp} found many tokens which were artifacts of the data used to pre-train the tokenizer, but presumably weren't present in the model's training data, leading to highly unpredictable (and often comical) completions. \citet{sun2023tokenization} find artifacts due to mismatches in tokenization in extractive Q\&A tasks, which may have connection to some of our experiments in Section~\ref{sec:repeat_output_control}. \citet{lundberg2023tokenhealing} propose \textit{token healing} to avoid many tokenization-related issues by removing the last few tokens from a prompt and allowing the model to complete them; this approach has connections to work on asking models to \textit{rephrase-and-respond} \cite{deng2023rephrase} and our experiments on prompting the model to repeat with its preferred tokenization direction (Section~\ref{sec:repeat_experiments}). Our work builds on these past scattered artifacts and provides a systematic analysis of tokenization-direction-dependent effects on numerical reasoning in frontier LLMs.

\textbf{Arithmetic tasks as a testbed for numerical reasoning in LLMs} With the increased interest in measuring frontier models on math reasoning \cite{saxton2018math, cobbe2021gsm8k, lewkowycz2022minerva, hendrycks2021math, paster2023hungarian}, an accompanying body of work studies language models in more controlled settings, such as arithmetic. \citet{razeghi2022frequencies} use arithmetic tasks to show that pre-training term frequencies\footnote{We also considered frequency effects by utilizing BPE merge ranks (given that we do not have access to the pre-training data of GPT-3.5 and GPT-4 models) as an approximate for frequency. We didn't find a strong effect, expanded results are provided in Appendix~\ref{appx:frequency}.} can affect numerical reasoning in GPT-J models \cite{wang2021gptj} trained on the Pile dataset \cite{gao2020pile}. Similarly, \citet{mccoy2023embers} showed that GPT-3.5 and GPT-4 are better at computing linear functions that are more common in training data (such as the Fahrenheit to Celsius conversion) than close alternatives. Other work \cite{nogueira2021arithmetic, muffo2022arithmetic, zhou2022algo_prompting, zhou2023rasp} instead focuses on various modifications that can help models generalize to longer arithmetic tasks. \citet{zhou2023rasp} and \citet{lee2023arithmetic} both point out that having autoregressive models perform addition in reversed order yields a simpler algorithm to learn and results in better performance, which is complementary to our emphasis on the importance of ``reversed'' (i.e. right-to-left) tokenization alignment. \citet{zhou2022algo_prompting} also conduct preliminary error analyses of model mistakes, though their algorithmic prompts force models to split tokens into digits (similar to \citet{nye2021scratchpad}). Our work broadly lies in this category of work using arithmetic tasks to study numerical reasoning; we chose to focus on tokenization-dependent effects, and found surprisingly consistent, stereotyped error patterns (Section~\ref{sec:error_patterns_overall}), adding to this rich body of literature. 

\section{Discussion}
\label{sec:discussion}

In this work, we analyze tokenization-dependent effects on numerical reasoning in GPT-3.5 and GPT-4. We found that the hard-coded choice of 1-, 2- and 3- digit number tokens, tokenized left-to-right, gives rise to stereotyped error patterns when compared to right-to-left tokenization. We proposed a mitigation, where the model is asked to repeat the answer in its preferred tokenization format. Finally, we showed that the effect is stronger in smaller models (such as GPT-3.5), emphasizing the significance of tokenization-dependent inductive biases in an era where many practitioners are focusing on packing capabilities into smaller models through overtraining \citep{devries2023overtraining, touvron2023llama1} and distillation \cite{li2023textbooks,geminiteam2023gemini}. Overall, we believe this evidence strongly suggests inductive biases from tokenization can significantly influence model performance on numerical reasoning tasks.

Modern frontier LLMs mostly use single-digit tokens (Table \ref{tab:tokenization_strategies}), with GPT-3.5 and GPT-4 being a key exception in their use of up-to-3-digit tokens. We hypothesize that the latter choice may have been made to achieve a better \textit{compression rate}: models ``see'' more numerical data for the same number of training tokens.\footnote{3-digit number tokens also reduce inference-time compute when models use numbers in their output, which could be an important consideration when serving models at scale.} Furthermore, this choice could have benefits for \textit{length generalization} \cite{anil2022exploring}, as we allude to in Section~\ref{sec:errors_rest}. However, we've also demonstrated how the misalignment between inputs and outputs when using L2R tokenization (Section~\ref{sec:error_length}) can lead to large drops in accuracy, especially on smaller models (GPT-3.5, GPT-4 Turbo). Such misalignment would not be an issue when using single-digit tokens.

To make progress on which number tokenization choices are best to use (e.g., the single-digit tokens of LLaMa and PaLM, or the up-to-3 digit tokens of GPT-3.5 and GPT-4), the ``gold experiment'' would be to train the same model architecture on the same dataset, but with varying number tokenization strategies. Beyond the expense of this experiment (making it intractable in academic settings), a key question also becomes how to ``compute''-control. The better compression ratio of up-to-3 digit tokens means a token-controlled experiment would result in some models ``seeing'' more data. We hope our work leads model practitioners to consider such ablations, with proper controls.

Beyond applicability to model practitioners, our work also provides an interesting set of tokenizaiton-dependent phenomenon for interpretability researchers to explore. Prior work \cite{stolfo2023mechanistic} has used techniques such as \textit{path patching} to identify sub-circuits in LLMs that perform arithmetic tasks, but restricted to single token operands. Building off our results, it would be interesting to elucidate the mechanisms behind systematic error patterns, especially in the case of multi-token operands. The robustness of the ``digit 4'' error on GPT-3.5 points to some systematic mechanism, which could shed light on underlying algorithms that emerge to perform arithmetic tasks.

\section*{Acknowledgements}

The authors would like to acknowledge Andrew Saxe, Ted Moskovitz, Kira Düsterwald, Felix Hill, Xavier Garcia, Dan Roberts, and William Held for insightful discussions and feedback on the draft. A.K.S. is funded by the Gatsby Charitable Foundation.

\nocite{touvron2023llama2}

\bibliography{references}

\begin{thebibliography}{45}
\providecommand{\natexlab}[1]{#1}
\providecommand{\url}[1]{\texttt{#1}}
\expandafter\ifx\csname urlstyle\endcsname\relax
  \providecommand{\doi}[1]{doi: #1}\else
  \providecommand{\doi}{doi: \begingroup \urlstyle{rm}\Url}\fi

\bibitem[Anil et~al.(2022)Anil, Wu, Andreassen, Lewkowycz, Misra, Ramasesh, Slone, Gur-Ari, Dyer, and Neyshabur]{anil2022exploring}
Anil, C., Wu, Y., Andreassen, A.~J., Lewkowycz, A., Misra, V., Ramasesh, V.~V., Slone, A., Gur-Ari, G., Dyer, E., and Neyshabur, B.
\newblock Exploring length generalization in large language models.
\newblock \emph{Neural Information Processing Systems (NeurIPS)}, 2022.
\newblock URL \url{https://openreview.net/forum?id=zSkYVeX7bC4}.

\bibitem[Bostrom \& Durrett(2020)Bostrom and Durrett]{bostrom2020bpe}
Bostrom, K. and Durrett, G.
\newblock Byte pair encoding is suboptimal for language model pretraining.
\newblock \emph{Empirical Methods in Natural Language Processing (EMNLP)}, 2020.
\newblock URL \url{https://aclanthology.org/2020.findings-emnlp.414}.

\bibitem[Brown et~al.(2020)Brown, Mann, Ryder, Subbiah, Kaplan, Dhariwal, Neelakantan, Shyam, Sastry, Askell, Agarwal, Herbert-Voss, Krueger, Henighan, Child, Ramesh, Ziegler, Wu, Winter, Hesse, Chen, Sigler, Litwin, Gray, Chess, Clark, Berner, McCandlish, Radford, Sutskever, and Amodei]{brown2020gpt3}
Brown, T., Mann, B., Ryder, N., Subbiah, M., Kaplan, J.~D., Dhariwal, P., Neelakantan, A., Shyam, P., Sastry, G., Askell, A., Agarwal, S., Herbert-Voss, A., Krueger, G., Henighan, T., Child, R., Ramesh, A., Ziegler, D., Wu, J., Winter, C., Hesse, C., Chen, M., Sigler, E., Litwin, M., Gray, S., Chess, B., Clark, J., Berner, C., McCandlish, S., Radford, A., Sutskever, I., and Amodei, D.
\newblock Language models are few-shot learners.
\newblock \emph{Neural Information Processing Systems (NeurIPS)}, 2020.
\newblock URL \url{https://proceedings.neurips.cc/paper_files/paper/2020/file/1457c0d6bfcb4967418bfb8ac142f64a-Paper.pdf}.

\bibitem[Chowdhery et~al.(2023)Chowdhery, Narang, Devlin, Bosma, Mishra, Roberts, Barham, Chung, Sutton, Gehrmann, Schuh, Shi, Tsvyashchenko, Maynez, Rao, Barnes, Tay, Shazeer, Prabhakaran, Reif, Du, Hutchinson, Pope, Bradbury, Austin, Isard, Gur-Ari, Yin, Duke, Levskaya, Ghemawat, Dev, Michalewski, Garcia, Misra, Robinson, Fedus, Zhou, Ippolito, Luan, Lim, Zoph, Spiridonov, Sepassi, Dohan, Agrawal, Omernick, Dai, Pillai, Pellat, Lewkowycz, Moreira, Child, Polozov, Lee, Zhou, Wang, Saeta, Diaz, Firat, Catasta, Wei, Meier-Hellstern, Eck, Dean, Petrov, and Fiedel]{chowdhery2022palm}
Chowdhery, A., Narang, S., Devlin, J., Bosma, M., Mishra, G., Roberts, A., Barham, P., Chung, H.~W., Sutton, C., Gehrmann, S., Schuh, P., Shi, K., Tsvyashchenko, S., Maynez, J., Rao, A., Barnes, P., Tay, Y., Shazeer, N., Prabhakaran, V., Reif, E., Du, N., Hutchinson, B., Pope, R., Bradbury, J., Austin, J., Isard, M., Gur-Ari, G., Yin, P., Duke, T., Levskaya, A., Ghemawat, S., Dev, S., Michalewski, H., Garcia, X., Misra, V., Robinson, K., Fedus, L., Zhou, D., Ippolito, D., Luan, D., Lim, H., Zoph, B., Spiridonov, A., Sepassi, R., Dohan, D., Agrawal, S., Omernick, M., Dai, A.~M., Pillai, T.~S., Pellat, M., Lewkowycz, A., Moreira, E., Child, R., Polozov, O., Lee, K., Zhou, Z., Wang, X., Saeta, B., Diaz, M., Firat, O., Catasta, M., Wei, J., Meier-Hellstern, K., Eck, D., Dean, J., Petrov, S., and Fiedel, N.
\newblock {PaLM}: Scaling language modeling with pathways.
\newblock \emph{Journal of Machine Learning Research}, 2023.
\newblock URL \url{http://jmlr.org/papers/v24/22-1144.html}.

\bibitem[Cobbe et~al.(2021)Cobbe, Kosaraju, Bavarian, Chen, Jun, Kaiser, Plappert, Tworek, Hilton, Nakano, et~al.]{cobbe2021gsm8k}
Cobbe, K., Kosaraju, V., Bavarian, M., Chen, M., Jun, H., Kaiser, L., Plappert, M., Tworek, J., Hilton, J., Nakano, R., et~al.
\newblock Training verifiers to solve math word problems.
\newblock \emph{arXiv:2110.14168}, 2021.
\newblock URL \url{https://arxiv.org/abs/2110.14168}.

\bibitem[De~Vries(2023)]{devries2023overtraining}
De~Vries, H.
\newblock Go smol or go home, 2023.
\newblock URL \url{https://www.harmdevries.com/post/model-size-vs-compute-overhead/}.

\bibitem[Deng et~al.(2023)Deng, Zhang, Chen, and Gu]{deng2023rephrase}
Deng, Y., Zhang, W., Chen, Z., and Gu, Q.
\newblock Rephrase and respond: Let large language models ask better questions for themselves.
\newblock \emph{arXiv:2311.04205}, 2023.
\newblock URL \url{https://arxiv.org/abs/2311.04205}.

\bibitem[Gage(1994)]{gage1994bpe}
Gage, P.
\newblock A new algorithm for data compression.
\newblock \emph{C Users Journal}, 1994.

\bibitem[Gao et~al.(2020)Gao, Biderman, Black, Golding, Hoppe, Foster, Phang, He, Thite, Nabeshima, et~al.]{gao2020pile}
Gao, L., Biderman, S., Black, S., Golding, L., Hoppe, T., Foster, C., Phang, J., He, H., Thite, A., Nabeshima, N., et~al.
\newblock {The Pile}: An 800gb dataset of diverse text for language modeling.
\newblock \emph{arXiv:2101.00027}, 2020.
\newblock URL \url{https://arxiv.org/abs/2101.00027}.

\bibitem[{Gemini Team} et~al.(2023){Gemini Team}, Anil, Borgeaud, Wu, Alayrac, Yu, Soricut, Schalkwyk, Dai, Hauth, Millican, Silver, Petrov, Johnson, Antonoglou, Schrittwieser, Glaese, Chen, Pitler, Lillicrap, Lazaridou, Firat, Molloy, Isard, Barham, Hennigan, Lee, Viola, Reynolds, Xu, Doherty, Collins, Meyer, Rutherford, Moreira, Ayoub, Goel, Tucker, Piqueras, Krikun, Barr, Savinov, Danihelka, Roelofs, White, Andreassen, von Glehn, Yagati, Kazemi, Gonzalez, Khalman, Sygnowski, Frechette, Smith, Culp, Proleev, Luan, Chen, Lottes, Schucher, Lebron, Rrustemi, Clay, Crone, Kocisky, Zhao, Perz, Yu, Howard, Bloniarz, Rae, Lu, Sifre, Maggioni, Alcober, Garrette, Barnes, Thakoor, Austin, Barth-Maron, Wong, Joshi, Chaabouni, Fatiha, Ahuja, Liu, Li, Cogan, Chen, Jia, Gu, Zhang, Grimstad, Hartman, Chadwick, Tomar, Garcia, Senter, Taropa, Pillai, Devlin, Laskin, de~Las~Casas, Valter, Tao, Blanco, Badia, Reitter, Chen, Brennan, Rivera, Brin, Iqbal, Surita, Labanowski, Rao, Winkler, Parisotto, Gu, Olszewska, Zhang,
  Addanki, Miech, Louis, Shafey, Teplyashin, Brown, Catt, Attaluri, Balaguer, Xiang, Wang, Ashwood, Briukhov, Webson, Ganapathy, Sanghavi, Kannan, Chang, Stjerngren, Djolonga, Sun, Bapna, Aitchison, Pejman, Michalewski, Yu, Wang, Love, Ahn, Bloxwich, Han, Humphreys, Sellam, Bradbury, Godbole, Samangooei, Damoc, Kaskasoli, Arnold, Vasudevan, Agrawal, Riesa, Lepikhin, Tanburn, Srinivasan, Lim, Hodkinson, Shyam, Ferret, Hand, Garg, Paine, Li, Li, Giang, Neitz, Abbas, York, Reid, Cole, Chowdhery, Das, Rogozińska, Nikolaev, Sprechmann, Nado, Zilka, Prost, He, Monteiro, Mishra, Welty, Newlan, Jia, Allamanis, Hu, de~Liedekerke, Gilmer, Saroufim, Rijhwani, Hou, Shrivastava, Baddepudi, Goldin, Ozturel, Cassirer, Xu, Sohn, Sachan, Amplayo, Swanson, Petrova, Narayan, Guez, Brahma, Landon, Patel, Zhao, Villela, Wang, Jia, Rahtz, Giménez, Yeung, Lin, Keeling, Georgiev, Mincu, Wu, Haykal, Saputro, Vodrahalli, Qin, Cankara, Sharma, Fernando, Hawkins, Neyshabur, Kim, Hutter, Agrawal, Castro-Ros, van~den Driessche, Wang,
  Yang, yiin Chang, Komarek, McIlroy, Lučić, Zhang, Farhan, Sharman, Natsev, Michel, Cheng, Bansal, Qiao, Cao, Shakeri, Butterfield, Chung, Rubenstein, Agrawal, Mensch, Soparkar, Lenc, Chung, Pope, Maggiore, Kay, Jhakra, Wang, Maynez, Phuong, Tobin, Tacchetti, Trebacz, Robinson, Katariya, Riedel, Bailey, Xiao, Ghelani, Aroyo, Slone, Houlsby, Xiong, Yang, Gribovskaya, Adler, Wirth, Lee, Li, Kagohara, Pavagadhi, Bridgers, Bortsova, Ghemawat, Ahmed, Liu, Powell, Bolina, Iinuma, Zablotskaia, Besley, Chung, Dozat, Comanescu, Si, Greer, Su, Polacek, Kaufman, Tokumine, Hu, Buchatskaya, Miao, Elhawaty, Siddhant, Tomasev, Xing, Greer, Miller, Ashraf, Roy, Zhang, Ma, Filos, Besta, Blevins, Klimenko, Yeh, Changpinyo, Mu, Chang, Pajarskas, Muir, Cohen, Lan, Haridasan, Marathe, Hansen, Douglas, Samuel, Wang, Austin, Lan, Jiang, Chiu, Lorenzo, Sjösund, Cevey, Gleicher, Avrahami, Boral, Srinivasan, Selo, May, Aisopos, Hussenot, Soares, Baumli, Chang, Recasens, Caine, Pritzel, Pavetic, Pardo, Gergely, Frye, Ramasesh,
  Horgan, Badola, Kassner, Roy, Dyer, Campos, Tomala, Tang, Badawy, White, Mustafa, Lang, Jindal, Vikram, Gong, Caelles, Hemsley, Thornton, Feng, Stokowiec, Zheng, Thacker, Çağlar Ünlü, Zhang, Saleh, Svensson, Bileschi, Patil, Anand, Ring, Tsihlas, Vezer, Selvi, Shevlane, Rodriguez, Kwiatkowski, Daruki, Rong, Dafoe, FitzGerald, Gu-Lemberg, Khan, Hendricks, Pellat, Feinberg, Cobon-Kerr, Sainath, Rauh, Hashemi, Ives, Hasson, Li, Noland, Cao, Byrd, Hou, Wang, Sottiaux, Paganini, Lespiau, Moufarek, Hassan, Shivakumar, van Amersfoort, Mandhane, Joshi, Goyal, Tung, Brock, Sheahan, Misra, Li, Rakićević, Dehghani, Liu, Mittal, Oh, Noury, Sezener, Huot, Lamm, Cao, Chen, Elsayed, Chi, Mahdieh, Tenney, Hua, Petrychenko, Kane, Scandinaro, Jain, Uesato, Datta, Sadovsky, Bunyan, Rabiej, Wu, Zhang, Vasudevan, Leurent, Alnahlawi, Georgescu, Wei, Zheng, Chan, Rabinovitch, Stanczyk, Zhang, Steiner, Naskar, Azzam, Johnson, Paszke, Chiu, Elias, Mohiuddin, Muhammad, Miao, Lee, Vieillard, Potluri, Park, Davoodi, Zhang,
  Stanway, Garmon, Karmarkar, Dong, Lee, Kumar, Zhou, Evens, Isaac, Chen, Jia, Levskaya, Zhu, Gorgolewski, Grabowski, Mao, Magni, Yao, Snaider, Casagrande, Suganthan, Palmer, Irving, Loper, Faruqui, Arkatkar, Chen, Shafran, Fink, Castaño, Giannoumis, Kim, Rybiński, Sreevatsa, Prendki, Soergel, Goedeckemeyer, Gierke, Jafari, Gaba, Wiesner, Wright, Wei, Vashisht, Kulizhskaya, Hoover, Le, Li, Iwuanyanwu, Liu, Ramirez, Khorlin, Cui, LIN, Georgiev, Wu, Aguilar, Pallo, Chakladar, Repina, Wu, van~der Weide, Ponnapalli, Kaplan, Simsa, Li, Dousse, Yang, Piper, Ie, Lui, Pasumarthi, Lintz, Vijayakumar, Thiet, Andor, Valenzuela, Paduraru, Peng, Lee, Zhang, Greene, Nguyen, Kurylowicz, Velury, Krause, Hardin, Dixon, Janzer, Choo, Feng, Zhang, Singhal, Latkar, Zhang, Le, Abellan, Du, McKinnon, Antropova, Bolukbasi, Keller, Reid, Finchelstein, Raad, Crocker, Hawkins, Dadashi, Gaffney, Lall, Franko, Filonov, Bulanova, Leblond, Yadav, Chung, Askham, Cobo, Xu, Fischer, Xu, Sorokin, Alberti, Lin, Evans, Zhou, Dimitriev,
  Forbes, Banarse, Tung, Liu, Omernick, Bishop, Kumar, Sterneck, Foley, Jain, Mishra, Xia, Bos, Cideron, Amid, Piccinno, Wang, Banzal, Gurita, Noga, Shah, Mankowitz, Polozov, Kushman, Krakovna, Brown, Bateni, Duan, Firoiu, Thotakuri, Natan, Mohananey, Geist, Mudgal, Girgin, Li, Ye, Roval, Tojo, Kwong, Lee-Thorp, Yew, Yuan, Bagri, Sinopalnikov, Ramos, Mellor, Sharma, Severyn, Lai, Wu, Cheng, Miller, Sonnerat, Vnukov, Greig, Beattie, Caveness, Bai, Eisenschlos, Korchemniy, Tsai, Jasarevic, Kong, Dao, Zheng, Liu, Yang, Zhu, Geller, Teh, Sanmiya, Gladchenko, Trdin, Sozanschi, Toyama, Rosen, Tavakkol, Xue, Elkind, Woodman, Carpenter, Papamakarios, Kemp, Kafle, Grunina, Sinha, Talbert, Goyal, Wu, Owusu-Afriyie, Du, Thornton, Pont-Tuset, Narayana, Li, Fatehi, Wieting, Ajmeri, Uria, Zhu, Ko, Knight, Héliou, Niu, Gu, Pang, Tran, Li, Levine, Stolovich, Kalb, Santamaria-Fernandez, Goenka, Yustalim, Strudel, Elqursh, Lakshminarayanan, Deck, Upadhyay, Lee, Dusenberry, Li, Wang, Levin, Hoffmann, Holtmann-Rice, Bachem,
  Yue, Arora, Malmi, Mirylenka, Tan, Koh, Yeganeh, Põder, Zheng, Pongetti, Tariq, Sun, Ionita, Seyedhosseini, Tafti, Kotikalapudi, Liu, Gulati, Liu, Ye, Chrzaszcz, Wang, Sethi, Li, Brown, Singh, Fan, Parisi, Stanton, Kuang, Koverkathu, Choquette-Choo, Li, Lu, Ittycheriah, Shroff, Sun, Varadarajan, Bahargam, Willoughby, Gaddy, Dasgupta, Desjardins, Cornero, Robenek, Mittal, Albrecht, Shenoy, Moiseev, Jacobsson, Ghaffarkhah, Rivière, Walton, Crepy, Parrish, Liu, Zhou, Farabet, Radebaugh, Srinivasan, van~der Salm, Fidjeland, Scellato, Latorre-Chimoto, Klimczak-Plucińska, Bridson, de~Cesare, Hudson, Mendolicchio, Walker, Morris, Penchev, Mauger, Guseynov, Reid, Odoom, Loher, Cotruta, Yenugula, Grewe, Petrushkina, Duerig, Sanchez, Yadlowsky, Shen, Globerson, Kurzrok, Webb, Dua, Li, Lahoti, Bhupatiraju, Hurt, Qureshi, Agarwal, Shani, Eyal, Khare, Belle, Wang, Tekur, Kale, Wei, Sang, Saeta, Liechty, Sun, Zhao, Lee, Nayak, Fritz, Vuyyuru, Aslanides, Vyas, Wicke, Ma, Bilal, Eltyshev, Balle, Martin, Cate, Manyika,
  Amiri, Kim, Xiong, Kang, Luisier, Tripuraneni, Madras, Guo, Waters, Wang, Ainslie, Baldridge, Zhang, Pruthi, Bauer, Yang, Mansour, Gelman, Xu, Polovets, Liu, Cai, Chen, Sheng, Xue, Ozair, Yu, Angermueller, Li, Wang, Wiesinger, Koukoumidis, Tian, Iyer, Gurumurthy, Goldenson, Shah, Blake, Yu, Urbanowicz, Palomaki, Fernando, Brooks, Durden, Mehta, Momchev, Rahimtoroghi, Georgaki, Raul, Ruder, Redshaw, Lee, Jalan, Li, Perng, Hechtman, Schuh, Nasr, Chen, Milan, Mikulik, Strohman, Franco, Green, Hassabis, Kavukcuoglu, Dean, and Vinyals]{geminiteam2023gemini}
{Gemini Team}, Anil, R., Borgeaud, S., Wu, Y., Alayrac, J.-B., Yu, J., Soricut, R., Schalkwyk, J., Dai, A.~M., Hauth, A., Millican, K., Silver, D., Petrov, S., Johnson, M., Antonoglou, I., Schrittwieser, J., Glaese, A., Chen, J., Pitler, E., Lillicrap, T., Lazaridou, A., Firat, O., Molloy, J., Isard, M., Barham, P.~R., Hennigan, T., Lee, B., Viola, F., Reynolds, M., Xu, Y., Doherty, R., Collins, E., Meyer, C., Rutherford, E., Moreira, E., Ayoub, K., Goel, M., Tucker, G., Piqueras, E., Krikun, M., Barr, I., Savinov, N., Danihelka, I., Roelofs, B., White, A., Andreassen, A., von Glehn, T., Yagati, L., Kazemi, M., Gonzalez, L., Khalman, M., Sygnowski, J., Frechette, A., Smith, C., Culp, L., Proleev, L., Luan, Y., Chen, X., Lottes, J., Schucher, N., Lebron, F., Rrustemi, A., Clay, N., Crone, P., Kocisky, T., Zhao, J., Perz, B., Yu, D., Howard, H., Bloniarz, A., Rae, J.~W., Lu, H., Sifre, L., Maggioni, M., Alcober, F., Garrette, D., Barnes, M., Thakoor, S., Austin, J., Barth-Maron, G., Wong, W., Joshi, R.,
  Chaabouni, R., Fatiha, D., Ahuja, A., Liu, R., Li, Y., Cogan, S., Chen, J., Jia, C., Gu, C., Zhang, Q., Grimstad, J., Hartman, A.~J., Chadwick, M., Tomar, G.~S., Garcia, X., Senter, E., Taropa, E., Pillai, T.~S., Devlin, J., Laskin, M., de~Las~Casas, D., Valter, D., Tao, C., Blanco, L., Badia, A.~P., Reitter, D., Chen, M., Brennan, J., Rivera, C., Brin, S., Iqbal, S., Surita, G., Labanowski, J., Rao, A., Winkler, S., Parisotto, E., Gu, Y., Olszewska, K., Zhang, Y., Addanki, R., Miech, A., Louis, A., Shafey, L.~E., Teplyashin, D., Brown, G., Catt, E., Attaluri, N., Balaguer, J., Xiang, J., Wang, P., Ashwood, Z., Briukhov, A., Webson, A., Ganapathy, S., Sanghavi, S., Kannan, A., Chang, M.-W., Stjerngren, A., Djolonga, J., Sun, Y., Bapna, A., Aitchison, M., Pejman, P., Michalewski, H., Yu, T., Wang, C., Love, J., Ahn, J., Bloxwich, D., Han, K., Humphreys, P., Sellam, T., Bradbury, J., Godbole, V., Samangooei, S., Damoc, B., Kaskasoli, A., Arnold, S. M.~R., Vasudevan, V., Agrawal, S., Riesa, J., Lepikhin, D.,
  Tanburn, R., Srinivasan, S., Lim, H., Hodkinson, S., Shyam, P., Ferret, J., Hand, S., Garg, A., Paine, T.~L., Li, J., Li, Y., Giang, M., Neitz, A., Abbas, Z., York, S., Reid, M., Cole, E., Chowdhery, A., Das, D., Rogozińska, D., Nikolaev, V., Sprechmann, P., Nado, Z., Zilka, L., Prost, F., He, L., Monteiro, M., Mishra, G., Welty, C., Newlan, J., Jia, D., Allamanis, M., Hu, C.~H., de~Liedekerke, R., Gilmer, J., Saroufim, C., Rijhwani, S., Hou, S., Shrivastava, D., Baddepudi, A., Goldin, A., Ozturel, A., Cassirer, A., Xu, Y., Sohn, D., Sachan, D., Amplayo, R.~K., Swanson, C., Petrova, D., Narayan, S., Guez, A., Brahma, S., Landon, J., Patel, M., Zhao, R., Villela, K., Wang, L., Jia, W., Rahtz, M., Giménez, M., Yeung, L., Lin, H., Keeling, J., Georgiev, P., Mincu, D., Wu, B., Haykal, S., Saputro, R., Vodrahalli, K., Qin, J., Cankara, Z., Sharma, A., Fernando, N., Hawkins, W., Neyshabur, B., Kim, S., Hutter, A., Agrawal, P., Castro-Ros, A., van~den Driessche, G., Wang, T., Yang, F., yiin Chang, S., Komarek,
  P., McIlroy, R., Lučić, M., Zhang, G., Farhan, W., Sharman, M., Natsev, P., Michel, P., Cheng, Y., Bansal, Y., Qiao, S., Cao, K., Shakeri, S., Butterfield, C., Chung, J., Rubenstein, P.~K., Agrawal, S., Mensch, A., Soparkar, K., Lenc, K., Chung, T., Pope, A., Maggiore, L., Kay, J., Jhakra, P., Wang, S., Maynez, J., Phuong, M., Tobin, T., Tacchetti, A., Trebacz, M., Robinson, K., Katariya, Y., Riedel, S., Bailey, P., Xiao, K., Ghelani, N., Aroyo, L., Slone, A., Houlsby, N., Xiong, X., Yang, Z., Gribovskaya, E., Adler, J., Wirth, M., Lee, L., Li, M., Kagohara, T., Pavagadhi, J., Bridgers, S., Bortsova, A., Ghemawat, S., Ahmed, Z., Liu, T., Powell, R., Bolina, V., Iinuma, M., Zablotskaia, P., Besley, J., Chung, D.-W., Dozat, T., Comanescu, R., Si, X., Greer, J., Su, G., Polacek, M., Kaufman, R.~L., Tokumine, S., Hu, H., Buchatskaya, E., Miao, Y., Elhawaty, M., Siddhant, A., Tomasev, N., Xing, J., Greer, C., Miller, H., Ashraf, S., Roy, A., Zhang, Z., Ma, A., Filos, A., Besta, M., Blevins, R., Klimenko, T.,
  Yeh, C.-K., Changpinyo, S., Mu, J., Chang, O., Pajarskas, M., Muir, C., Cohen, V., Lan, C.~L., Haridasan, K., Marathe, A., Hansen, S., Douglas, S., Samuel, R., Wang, M., Austin, S., Lan, C., Jiang, J., Chiu, J., Lorenzo, J.~A., Sjösund, L.~L., Cevey, S., Gleicher, Z., Avrahami, T., Boral, A., Srinivasan, H., Selo, V., May, R., Aisopos, K., Hussenot, L., Soares, L.~B., Baumli, K., Chang, M.~B., Recasens, A., Caine, B., Pritzel, A., Pavetic, F., Pardo, F., Gergely, A., Frye, J., Ramasesh, V., Horgan, D., Badola, K., Kassner, N., Roy, S., Dyer, E., Campos, V., Tomala, A., Tang, Y., Badawy, D.~E., White, E., Mustafa, B., Lang, O., Jindal, A., Vikram, S., Gong, Z., Caelles, S., Hemsley, R., Thornton, G., Feng, F., Stokowiec, W., Zheng, C., Thacker, P., Çağlar Ünlü, Zhang, Z., Saleh, M., Svensson, J., Bileschi, M., Patil, P., Anand, A., Ring, R., Tsihlas, K., Vezer, A., Selvi, M., Shevlane, T., Rodriguez, M., Kwiatkowski, T., Daruki, S., Rong, K., Dafoe, A., FitzGerald, N., Gu-Lemberg, K., Khan, M.,
  Hendricks, L.~A., Pellat, M., Feinberg, V., Cobon-Kerr, J., Sainath, T., Rauh, M., Hashemi, S.~H., Ives, R., Hasson, Y., Li, Y., Noland, E., Cao, Y., Byrd, N., Hou, L., Wang, Q., Sottiaux, T., Paganini, M., Lespiau, J.-B., Moufarek, A., Hassan, S., Shivakumar, K., van Amersfoort, J., Mandhane, A., Joshi, P., Goyal, A., Tung, M., Brock, A., Sheahan, H., Misra, V., Li, C., Rakićević, N., Dehghani, M., Liu, F., Mittal, S., Oh, J., Noury, S., Sezener, E., Huot, F., Lamm, M., Cao, N.~D., Chen, C., Elsayed, G., Chi, E., Mahdieh, M., Tenney, I., Hua, N., Petrychenko, I., Kane, P., Scandinaro, D., Jain, R., Uesato, J., Datta, R., Sadovsky, A., Bunyan, O., Rabiej, D., Wu, S., Zhang, J., Vasudevan, G., Leurent, E., Alnahlawi, M., Georgescu, I., Wei, N., Zheng, I., Chan, B., Rabinovitch, P.~G., Stanczyk, P., Zhang, Y., Steiner, D., Naskar, S., Azzam, M., Johnson, M., Paszke, A., Chiu, C.-C., Elias, J.~S., Mohiuddin, A., Muhammad, F., Miao, J., Lee, A., Vieillard, N., Potluri, S., Park, J., Davoodi, E., Zhang, J.,
  Stanway, J., Garmon, D., Karmarkar, A., Dong, Z., Lee, J., Kumar, A., Zhou, L., Evens, J., Isaac, W., Chen, Z., Jia, J., Levskaya, A., Zhu, Z., Gorgolewski, C., Grabowski, P., Mao, Y., Magni, A., Yao, K., Snaider, J., Casagrande, N., Suganthan, P., Palmer, E., Irving, G., Loper, E., Faruqui, M., Arkatkar, I., Chen, N., Shafran, I., Fink, M., Castaño, A., Giannoumis, I., Kim, W., Rybiński, M., Sreevatsa, A., Prendki, J., Soergel, D., Goedeckemeyer, A., Gierke, W., Jafari, M., Gaba, M., Wiesner, J., Wright, D.~G., Wei, Y., Vashisht, H., Kulizhskaya, Y., Hoover, J., Le, M., Li, L., Iwuanyanwu, C., Liu, L., Ramirez, K., Khorlin, A., Cui, A., LIN, T., Georgiev, M., Wu, M., Aguilar, R., Pallo, K., Chakladar, A., Repina, A., Wu, X., van~der Weide, T., Ponnapalli, P., Kaplan, C., Simsa, J., Li, S., Dousse, O., Yang, F., Piper, J., Ie, N., Lui, M., Pasumarthi, R., Lintz, N., Vijayakumar, A., Thiet, L.~N., Andor, D., Valenzuela, P., Paduraru, C., Peng, D., Lee, K., Zhang, S., Greene, S., Nguyen, D.~D., Kurylowicz,
  P., Velury, S., Krause, S., Hardin, C., Dixon, L., Janzer, L., Choo, K., Feng, Z., Zhang, B., Singhal, A., Latkar, T., Zhang, M., Le, Q., Abellan, E.~A., Du, D., McKinnon, D., Antropova, N., Bolukbasi, T., Keller, O., Reid, D., Finchelstein, D., Raad, M.~A., Crocker, R., Hawkins, P., Dadashi, R., Gaffney, C., Lall, S., Franko, K., Filonov, E., Bulanova, A., Leblond, R., Yadav, V., Chung, S., Askham, H., Cobo, L.~C., Xu, K., Fischer, F., Xu, J., Sorokin, C., Alberti, C., Lin, C.-C., Evans, C., Zhou, H., Dimitriev, A., Forbes, H., Banarse, D., Tung, Z., Liu, J., Omernick, M., Bishop, C., Kumar, C., Sterneck, R., Foley, R., Jain, R., Mishra, S., Xia, J., Bos, T., Cideron, G., Amid, E., Piccinno, F., Wang, X., Banzal, P., Gurita, P., Noga, H., Shah, P., Mankowitz, D.~J., Polozov, A., Kushman, N., Krakovna, V., Brown, S., Bateni, M., Duan, D., Firoiu, V., Thotakuri, M., Natan, T., Mohananey, A., Geist, M., Mudgal, S., Girgin, S., Li, H., Ye, J., Roval, O., Tojo, R., Kwong, M., Lee-Thorp, J., Yew, C., Yuan, Q.,
  Bagri, S., Sinopalnikov, D., Ramos, S., Mellor, J., Sharma, A., Severyn, A., Lai, J., Wu, K., Cheng, H.-T., Miller, D., Sonnerat, N., Vnukov, D., Greig, R., Beattie, J., Caveness, E., Bai, L., Eisenschlos, J., Korchemniy, A., Tsai, T., Jasarevic, M., Kong, W., Dao, P., Zheng, Z., Liu, F., Yang, F., Zhu, R., Geller, M., Teh, T.~H., Sanmiya, J., Gladchenko, E., Trdin, N., Sozanschi, A., Toyama, D., Rosen, E., Tavakkol, S., Xue, L., Elkind, C., Woodman, O., Carpenter, J., Papamakarios, G., Kemp, R., Kafle, S., Grunina, T., Sinha, R., Talbert, A., Goyal, A., Wu, D., Owusu-Afriyie, D., Du, C., Thornton, C., Pont-Tuset, J., Narayana, P., Li, J., Fatehi, S., Wieting, J., Ajmeri, O., Uria, B., Zhu, T., Ko, Y., Knight, L., Héliou, A., Niu, N., Gu, S., Pang, C., Tran, D., Li, Y., Levine, N., Stolovich, A., Kalb, N., Santamaria-Fernandez, R., Goenka, S., Yustalim, W., Strudel, R., Elqursh, A., Lakshminarayanan, B., Deck, C., Upadhyay, S., Lee, H., Dusenberry, M., Li, Z., Wang, X., Levin, K., Hoffmann, R.,
  Holtmann-Rice, D., Bachem, O., Yue, S., Arora, S., Malmi, E., Mirylenka, D., Tan, Q., Koh, C., Yeganeh, S.~H., Põder, S., Zheng, S., Pongetti, F., Tariq, M., Sun, Y., Ionita, L., Seyedhosseini, M., Tafti, P., Kotikalapudi, R., Liu, Z., Gulati, A., Liu, J., Ye, X., Chrzaszcz, B., Wang, L., Sethi, N., Li, T., Brown, B., Singh, S., Fan, W., Parisi, A., Stanton, J., Kuang, C., Koverkathu, V., Choquette-Choo, C.~A., Li, Y., Lu, T., Ittycheriah, A., Shroff, P., Sun, P., Varadarajan, M., Bahargam, S., Willoughby, R., Gaddy, D., Dasgupta, I., Desjardins, G., Cornero, M., Robenek, B., Mittal, B., Albrecht, B., Shenoy, A., Moiseev, F., Jacobsson, H., Ghaffarkhah, A., Rivière, M., Walton, A., Crepy, C., Parrish, A., Liu, Y., Zhou, Z., Farabet, C., Radebaugh, C., Srinivasan, P., van~der Salm, C., Fidjeland, A., Scellato, S., Latorre-Chimoto, E., Klimczak-Plucińska, H., Bridson, D., de~Cesare, D., Hudson, T., Mendolicchio, P., Walker, L., Morris, A., Penchev, I., Mauger, M., Guseynov, A., Reid, A., Odoom, S., Loher,
  L., Cotruta, V., Yenugula, M., Grewe, D., Petrushkina, A., Duerig, T., Sanchez, A., Yadlowsky, S., Shen, A., Globerson, A., Kurzrok, A., Webb, L., Dua, S., Li, D., Lahoti, P., Bhupatiraju, S., Hurt, D., Qureshi, H., Agarwal, A., Shani, T., Eyal, M., Khare, A., Belle, S.~R., Wang, L., Tekur, C., Kale, M.~S., Wei, J., Sang, R., Saeta, B., Liechty, T., Sun, Y., Zhao, Y., Lee, S., Nayak, P., Fritz, D., Vuyyuru, M.~R., Aslanides, J., Vyas, N., Wicke, M., Ma, X., Bilal, T., Eltyshev, E., Balle, D., Martin, N., Cate, H., Manyika, J., Amiri, K., Kim, Y., Xiong, X., Kang, K., Luisier, F., Tripuraneni, N., Madras, D., Guo, M., Waters, A., Wang, O., Ainslie, J., Baldridge, J., Zhang, H., Pruthi, G., Bauer, J., Yang, F., Mansour, R., Gelman, J., Xu, Y., Polovets, G., Liu, J., Cai, H., Chen, W., Sheng, X., Xue, E., Ozair, S., Yu, A., Angermueller, C., Li, X., Wang, W., Wiesinger, J., Koukoumidis, E., Tian, Y., Iyer, A., Gurumurthy, M., Goldenson, M., Shah, P., Blake, M., Yu, H., Urbanowicz, A., Palomaki, J., Fernando,
  C., Brooks, K., Durden, K., Mehta, H., Momchev, N., Rahimtoroghi, E., Georgaki, M., Raul, A., Ruder, S., Redshaw, M., Lee, J., Jalan, K., Li, D., Perng, G., Hechtman, B., Schuh, P., Nasr, M., Chen, M., Milan, K., Mikulik, V., Strohman, T., Franco, J., Green, T., Hassabis, D., Kavukcuoglu, K., Dean, J., and Vinyals, O.
\newblock Gemini: A family of highly capable multimodal models, 2023.
\newblock URL \url{https://arxiv.org/abs/2312.11805}.

\bibitem[Golkar et~al.(2023)Golkar, Pettee, Eickenberg, Bietti, Cranmer, Krawezik, Lanusse, McCabe, Ohana, Parker, Blancard, Tesileanu, Cho, and Ho]{golkar2023xval}
Golkar, S., Pettee, M., Eickenberg, M., Bietti, A., Cranmer, M., Krawezik, G., Lanusse, F., McCabe, M., Ohana, R., Parker, L., Blancard, B. R.-S., Tesileanu, T., Cho, K., and Ho, S.
\newblock xval: A continuous number encoding for large language models.
\newblock \emph{Neural Information Processing Systems (NeurIPS) AI for Science Workshop}, 2023.
\newblock URL \url{https://openreview.net/forum?id=KHDMZtoF4i}.

\bibitem[Goyal et~al.(2024)Goyal, Ji, Rawat, Menon, Kumar, and Nagarajan]{goyal2024pause}
Goyal, S., Ji, Z., Rawat, A.~S., Menon, A.~K., Kumar, S., and Nagarajan, V.
\newblock Think before you speak: Training language models with pause tokens.
\newblock \emph{International Conference on Learning Representations (ICLR)}, 2024.
\newblock URL \url{https://openreview.net/forum?id=ph04CRkPdC}.

\bibitem[Groeneveld et~al.(2024)Groeneveld, Beltagy, Walsh, Bhagia, Kinney, Tafjord, Jha, Ivison, Magnusson, Wang, Arora, Atkinson, Authur, Chandu, Cohan, Dumas, Elazar, Gu, Hessel, Khot, Merrill, Morrison, Muennighoff, Naik, Nam, Peters, Pyatkin, Ravichander, Schwenk, Shah, Smith, Strubell, Subramani, Wortsman, Dasigi, Lambert, Richardson, Zettlemoyer, Dodge, Lo, Soldaini, Smith, and Hajishirzi]{groeneveld2024olmo}
Groeneveld, D., Beltagy, I., Walsh, P., Bhagia, A., Kinney, R., Tafjord, O., Jha, A.~H., Ivison, H., Magnusson, I., Wang, Y., Arora, S., Atkinson, D., Authur, R., Chandu, K.~R., Cohan, A., Dumas, J., Elazar, Y., Gu, Y., Hessel, J., Khot, T., Merrill, W., Morrison, J., Muennighoff, N., Naik, A., Nam, C., Peters, M.~E., Pyatkin, V., Ravichander, A., Schwenk, D., Shah, S., Smith, W., Strubell, E., Subramani, N., Wortsman, M., Dasigi, P., Lambert, N., Richardson, K., Zettlemoyer, L., Dodge, J., Lo, K., Soldaini, L., Smith, N.~A., and Hajishirzi, H.
\newblock Olmo: Accelerating the science of language models, 2024.

\bibitem[Hendrycks et~al.(2021)Hendrycks, Burns, Kadavath, Arora, Basart, Tang, Song, and Steinhardt]{hendrycks2021math}
Hendrycks, D., Burns, C., Kadavath, S., Arora, A., Basart, S., Tang, E., Song, D., and Steinhardt, J.
\newblock Measuring mathematical problem solving with the {MATH} dataset.
\newblock \emph{Neural Information Processing Systems (NeurIPS) Datasets and Benchmarks Track}, 2021.
\newblock URL \url{https://openreview.net/forum?id=7Bywt2mQsCe}.

\bibitem[Jiang et~al.(2023)Jiang, Sablayrolles, Mensch, Bamford, Chaplot, de~las Casas, Bressand, Lengyel, Lample, Saulnier, Lavaud, Lachaux, Stock, Scao, Lavril, Wang, Lacroix, and Sayed]{jiang2023mistral}
Jiang, A.~Q., Sablayrolles, A., Mensch, A., Bamford, C., Chaplot, D.~S., de~las Casas, D., Bressand, F., Lengyel, G., Lample, G., Saulnier, L., Lavaud, L.~R., Lachaux, M.-A., Stock, P., Scao, T.~L., Lavril, T., Wang, T., Lacroix, T., and Sayed, W.~E.
\newblock Mistral 7b.
\newblock \emph{arXiv:2310.06825}, 2023.
\newblock URL \url{https://arxiv.org/abs/2310.06825}.

\bibitem[Kojima et~al.(2022)Kojima, Gu, Reid, Matsuo, and Iwasawa]{kojima2022cot}
Kojima, T., Gu, S.~S., Reid, M., Matsuo, Y., and Iwasawa, Y.
\newblock Large language models are zero-shot reasoners.
\newblock \emph{Neural Information Processing Systems (NeurIPS)}, 2022.
\newblock URL \url{https://openreview.net/forum?id=e2TBb5y0yFf}.

\bibitem[Kudo(2018)]{kudo2018unigram}
Kudo, T.
\newblock Subword regularization: Improving neural network translation models with multiple subword candidates.
\newblock \emph{Association for Computational Linguistics (ACL)}, 2018.
\newblock URL \url{https://aclanthology.org/P18-1007}.

\bibitem[Lanham et~al.(2023)Lanham, Chen, Radhakrishnan, Steiner, Denison, Hernandez, Li, Durmus, Hubinger, Kernion, et~al.]{lanham2023faithfulness}
Lanham, T., Chen, A., Radhakrishnan, A., Steiner, B., Denison, C., Hernandez, D., Li, D., Durmus, E., Hubinger, E., Kernion, J., et~al.
\newblock Measuring faithfulness in chain-of-thought reasoning.
\newblock \emph{arXiv:2307.13702}, 2023.
\newblock URL \url{https://arxiv.org/abs/2307.13702}.

\bibitem[Lee et~al.(2023)Lee, Sreenivasan, Lee, Lee, and Papailiopoulos]{lee2023arithmetic}
Lee, N., Sreenivasan, K., Lee, J.~D., Lee, K., and Papailiopoulos, D.
\newblock Teaching arithmetic to small transformers.
\newblock \emph{arXiv:2307.03381}, 2023.
\newblock URL \url{https://arxiv.org/abs/2307.03381}.

\bibitem[Lewkowycz et~al.(2022)Lewkowycz, Andreassen, Dohan, Dyer, Michalewski, Ramasesh, Slone, Anil, Schlag, Gutman-Solo, Wu, Neyshabur, Gur-Ari, and Misra]{lewkowycz2022minerva}
Lewkowycz, A., Andreassen, A.~J., Dohan, D., Dyer, E., Michalewski, H., Ramasesh, V.~V., Slone, A., Anil, C., Schlag, I., Gutman-Solo, T., Wu, Y., Neyshabur, B., Gur-Ari, G., and Misra, V.
\newblock Solving quantitative reasoning problems with language models.
\newblock \emph{Neural Information Processing Systems (NeurIPS)}, 2022.
\newblock URL \url{https://openreview.net/forum?id=IFXTZERXdM7}.

\bibitem[Li et~al.(2023)Li, Bubeck, Eldan, Del~Giorno, Gunasekar, and Lee]{li2023textbooks}
Li, Y., Bubeck, S., Eldan, R., Del~Giorno, A., Gunasekar, S., and Lee, Y.~T.
\newblock {Textbooks Are All You Need II}: phi-1.5 technical report.
\newblock \emph{arXiv preprint arXiv:2309.05463}, 2023.

\bibitem[Liu et~al.(2023)Liu, Garrette, Saharia, Chan, Roberts, Narang, Blok, Mical, Norouzi, and Constant]{liu2023character}
Liu, R., Garrette, D., Saharia, C., Chan, W., Roberts, A., Narang, S., Blok, I., Mical, R., Norouzi, M., and Constant, N.
\newblock Character-aware models improve visual text rendering.
\newblock \emph{Association for Computational Linguistics (ACL)}, 2023.
\newblock URL \url{https://aclanthology.org/2023.acl-long.900}.

\bibitem[Lundberg(2023)]{lundberg2023tokenhealing}
Lundberg, S.
\newblock The art of prompt design: Prompt boundaries and token healing, 2023.
\newblock URL \url{https://towardsdatascience.com/the-art-of-prompt-design-prompt-boundaries-and-token-healing-3b2448b0be38}.

\bibitem[McCoy et~al.(2023)McCoy, Yao, Friedman, Hardy, and Griffiths]{mccoy2023embers}
McCoy, R.~T., Yao, S., Friedman, D., Hardy, M., and Griffiths, T.~L.
\newblock Embers of autoregression: Understanding large language models through the problem they are trained to solve.
\newblock \emph{arXiv:2309.13638}, 2023.
\newblock URL \url{https://arxiv.org/abs/2309.13638}.

\bibitem[Muffo et~al.(2022)Muffo, Cocco, and Bertino]{muffo2022arithmetic}
Muffo, M., Cocco, A., and Bertino, E.
\newblock Evaluating transformer language models on arithmetic operations using number decomposition.
\newblock \emph{Language Resources and Evaluation Conference (LREC)}, 2022.
\newblock URL \url{https://aclanthology.org/2022.lrec-1.30}.

\bibitem[Nogueira et~al.(2021)Nogueira, Jiang, and Lin]{nogueira2021arithmetic}
Nogueira, R., Jiang, Z., and Lin, J.
\newblock Investigating the limitations of transformers with simple arithmetic tasks.
\newblock \emph{arXiv:2102.13019}, 2021.
\newblock URL \url{https://arxiv.org/abs/2102.13019}.

\bibitem[Nye et~al.(2021)Nye, Andreassen, Gur-Ari, Michalewski, Austin, Bieber, Dohan, Lewkowycz, Bosma, Luan, et~al.]{nye2021scratchpad}
Nye, M., Andreassen, A.~J., Gur-Ari, G., Michalewski, H., Austin, J., Bieber, D., Dohan, D., Lewkowycz, A., Bosma, M., Luan, D., et~al.
\newblock Show your work: Scratchpads for intermediate computation with language models.
\newblock \emph{arXiv:2112.00114}, 2021.
\newblock URL \url{https://arxiv.org/abs/2112.00114}.

\bibitem[OpenAI et~al.(2023)OpenAI, Achiam, Adler, Agarwal, Ahmad, Akkaya, Aleman, Almeida, Altenschmidt, Altman, Anadkat, Avila, Babuschkin, Balaji, Balcom, Baltescu, Bao, Bavarian, Belgum, Bello, Berdine, Bernadett-Shapiro, Berner, Bogdonoff, Boiko, Boyd, Brakman, Brockman, Brooks, Brundage, Button, Cai, Campbell, Cann, Carey, Carlson, Carmichael, Chan, Chang, Chantzis, Chen, Chen, Chen, Chen, Chen, Chess, Cho, Chu, Chung, Cummings, Currier, Dai, Decareaux, Degry, Deutsch, Deville, Dhar, Dohan, Dowling, Dunning, Ecoffet, Eleti, Eloundou, Farhi, Fedus, Felix, Fishman, Forte, Fulford, Gao, Georges, Gibson, Goel, Gogineni, Goh, Gontijo-Lopes, Gordon, Grafstein, Gray, Greene, Gross, Gu, Guo, Hallacy, Han, Harris, He, Heaton, Heidecke, Hesse, Hickey, Hickey, Hoeschele, Houghton, Hsu, Hu, Hu, Huizinga, Jain, Jain, Jang, Jiang, Jiang, Jin, Jin, Jomoto, Jonn, Jun, Kaftan, Łukasz Kaiser, Kamali, Kanitscheider, Keskar, Khan, Kilpatrick, Kim, Kim, Kim, Kirchner, Kiros, Knight, Kokotajlo, Łukasz Kondraciuk, Kondrich,
  Konstantinidis, Kosic, Krueger, Kuo, Lampe, Lan, Lee, Leike, Leung, Levy, Li, Lim, Lin, Lin, Litwin, Lopez, Lowe, Lue, Makanju, Malfacini, Manning, Markov, Markovski, Martin, Mayer, Mayne, McGrew, McKinney, McLeavey, McMillan, McNeil, Medina, Mehta, Menick, Metz, Mishchenko, Mishkin, Monaco, Morikawa, Mossing, Mu, Murati, Murk, Mély, Nair, Nakano, Nayak, Neelakantan, Ngo, Noh, Ouyang, O'Keefe, Pachocki, Paino, Palermo, Pantuliano, Parascandolo, Parish, Parparita, Passos, Pavlov, Peng, Perelman, de~Avila Belbute~Peres, Petrov, de~Oliveira~Pinto, Michael, Pokorny, Pokrass, Pong, Powell, Power, Power, Proehl, Puri, Radford, Rae, Ramesh, Raymond, Real, Rimbach, Ross, Rotsted, Roussez, Ryder, Saltarelli, Sanders, Santurkar, Sastry, Schmidt, Schnurr, Schulman, Selsam, Sheppard, Sherbakov, Shieh, Shoker, Shyam, Sidor, Sigler, Simens, Sitkin, Slama, Sohl, Sokolowsky, Song, Staudacher, Such, Summers, Sutskever, Tang, Tezak, Thompson, Tillet, Tootoonchian, Tseng, Tuggle, Turley, Tworek, Uribe, Vallone, Vijayvergiya,
  Voss, Wainwright, Wang, Wang, Wang, Ward, Wei, Weinmann, Welihinda, Welinder, Weng, Weng, Wiethoff, Willner, Winter, Wolrich, Wong, Workman, Wu, Wu, Wu, Xiao, Xu, Yoo, Yu, Yuan, Zaremba, Zellers, Zhang, Zhang, Zhao, Zheng, Zhuang, Zhuk, and Zoph]{openai2023gpt4}
OpenAI, Achiam, J., Adler, S., Agarwal, S., Ahmad, L., Akkaya, I., Aleman, F.~L., Almeida, D., Altenschmidt, J., Altman, S., Anadkat, S., Avila, R., Babuschkin, I., Balaji, S., Balcom, V., Baltescu, P., Bao, H., Bavarian, M., Belgum, J., Bello, I., Berdine, J., Bernadett-Shapiro, G., Berner, C., Bogdonoff, L., Boiko, O., Boyd, M., Brakman, A.-L., Brockman, G., Brooks, T., Brundage, M., Button, K., Cai, T., Campbell, R., Cann, A., Carey, B., Carlson, C., Carmichael, R., Chan, B., Chang, C., Chantzis, F., Chen, D., Chen, S., Chen, R., Chen, J., Chen, M., Chess, B., Cho, C., Chu, C., Chung, H.~W., Cummings, D., Currier, J., Dai, Y., Decareaux, C., Degry, T., Deutsch, N., Deville, D., Dhar, A., Dohan, D., Dowling, S., Dunning, S., Ecoffet, A., Eleti, A., Eloundou, T., Farhi, D., Fedus, L., Felix, N., Fishman, S.~P., Forte, J., Fulford, I., Gao, L., Georges, E., Gibson, C., Goel, V., Gogineni, T., Goh, G., Gontijo-Lopes, R., Gordon, J., Grafstein, M., Gray, S., Greene, R., Gross, J., Gu, S.~S., Guo, Y., Hallacy,
  C., Han, J., Harris, J., He, Y., Heaton, M., Heidecke, J., Hesse, C., Hickey, A., Hickey, W., Hoeschele, P., Houghton, B., Hsu, K., Hu, S., Hu, X., Huizinga, J., Jain, S., Jain, S., Jang, J., Jiang, A., Jiang, R., Jin, H., Jin, D., Jomoto, S., Jonn, B., Jun, H., Kaftan, T., Łukasz Kaiser, Kamali, A., Kanitscheider, I., Keskar, N.~S., Khan, T., Kilpatrick, L., Kim, J.~W., Kim, C., Kim, Y., Kirchner, H., Kiros, J., Knight, M., Kokotajlo, D., Łukasz Kondraciuk, Kondrich, A., Konstantinidis, A., Kosic, K., Krueger, G., Kuo, V., Lampe, M., Lan, I., Lee, T., Leike, J., Leung, J., Levy, D., Li, C.~M., Lim, R., Lin, M., Lin, S., Litwin, M., Lopez, T., Lowe, R., Lue, P., Makanju, A., Malfacini, K., Manning, S., Markov, T., Markovski, Y., Martin, B., Mayer, K., Mayne, A., McGrew, B., McKinney, S.~M., McLeavey, C., McMillan, P., McNeil, J., Medina, D., Mehta, A., Menick, J., Metz, L., Mishchenko, A., Mishkin, P., Monaco, V., Morikawa, E., Mossing, D., Mu, T., Murati, M., Murk, O., Mély, D., Nair, A., Nakano, R.,
  Nayak, R., Neelakantan, A., Ngo, R., Noh, H., Ouyang, L., O'Keefe, C., Pachocki, J., Paino, A., Palermo, J., Pantuliano, A., Parascandolo, G., Parish, J., Parparita, E., Passos, A., Pavlov, M., Peng, A., Perelman, A., de~Avila Belbute~Peres, F., Petrov, M., de~Oliveira~Pinto, H.~P., Michael, Pokorny, Pokrass, M., Pong, V., Powell, T., Power, A., Power, B., Proehl, E., Puri, R., Radford, A., Rae, J., Ramesh, A., Raymond, C., Real, F., Rimbach, K., Ross, C., Rotsted, B., Roussez, H., Ryder, N., Saltarelli, M., Sanders, T., Santurkar, S., Sastry, G., Schmidt, H., Schnurr, D., Schulman, J., Selsam, D., Sheppard, K., Sherbakov, T., Shieh, J., Shoker, S., Shyam, P., Sidor, S., Sigler, E., Simens, M., Sitkin, J., Slama, K., Sohl, I., Sokolowsky, B., Song, Y., Staudacher, N., Such, F.~P., Summers, N., Sutskever, I., Tang, J., Tezak, N., Thompson, M., Tillet, P., Tootoonchian, A., Tseng, E., Tuggle, P., Turley, N., Tworek, J., Uribe, J. F.~C., Vallone, A., Vijayvergiya, A., Voss, C., Wainwright, C., Wang, J.~J.,
  Wang, A., Wang, B., Ward, J., Wei, J., Weinmann, C., Welihinda, A., Welinder, P., Weng, J., Weng, L., Wiethoff, M., Willner, D., Winter, C., Wolrich, S., Wong, H., Workman, L., Wu, S., Wu, J., Wu, M., Xiao, K., Xu, T., Yoo, S., Yu, K., Yuan, Q., Zaremba, W., Zellers, R., Zhang, C., Zhang, M., Zhao, S., Zheng, T., Zhuang, J., Zhuk, W., and Zoph, B.
\newblock {GPT-4 Technical Report}, 2023.
\newblock URL \url{https://arxiv.org/abs/2303.08774}.

\bibitem[Paster(2023)]{paster2023hungarian}
Paster, K.
\newblock Testing language models on a held-out high school national finals exam.
\newblock \url{https://huggingface.co/datasets/keirp/hungarian_national_hs_finals_exam}, 2023.

\bibitem[Razeghi et~al.(2022)Razeghi, Logan~IV, Gardner, and Singh]{razeghi2022frequencies}
Razeghi, Y., Logan~IV, R.~L., Gardner, M., and Singh, S.
\newblock Impact of pretraining term frequencies on few-shot numerical reasoning.
\newblock \emph{Empirical Methods in Natural Language Processing (EMNLP)}, 2022.
\newblock URL \url{https://aclanthology.org/2022.findings-emnlp.59}.

\bibitem[Rumbelow \& mwatkins(2023)Rumbelow and mwatkins]{rumbelow2023solidgoldmagikarp}
Rumbelow, J. and mwatkins.
\newblock Solidgoldmagikarp (plus, prompt generation), 2023.
\newblock URL \url{https://www.lesswrong.com/posts/aPeJE8bSo6rAFoLqg/solidgoldmagikarp-plus-prompt-generation}.

\bibitem[Saxton et~al.(2019)Saxton, Grefenstette, Hill, and Kohli]{saxton2018math}
Saxton, D., Grefenstette, E., Hill, F., and Kohli, P.
\newblock Analysing mathematical reasoning abilities of neural models.
\newblock \emph{International Conference on Learning Representations (ICLR)}, 2019.
\newblock URL \url{https://openreview.net/forum?id=H1gR5iR5FX}.

\bibitem[Sennrich et~al.(2016)Sennrich, Haddow, and Birch]{sennrich2016bpe}
Sennrich, R., Haddow, B., and Birch, A.
\newblock Neural machine translation of rare words with subword units.
\newblock \emph{Association for Computational Linguistics (ACL)}, 2016.
\newblock URL \url{https://aclanthology.org/P16-1162}.

\bibitem[Shin et~al.(2020)Shin, Razeghi, Logan~IV, Wallace, and Singh]{shin2020autoprompt}
Shin, T., Razeghi, Y., Logan~IV, R.~L., Wallace, E., and Singh, S.
\newblock {A}uto{P}rompt: {E}liciting {K}nowledge from {L}anguage {M}odels with {A}utomatically {G}enerated {P}rompts.
\newblock \emph{Empirical Methods in Natural Language Processing (EMNLP)}, 2020.
\newblock URL \url{https://aclanthology.org/2020.emnlp-main.346}.

\bibitem[Stolfo et~al.(2023)Stolfo, Belinkov, and Sachan]{stolfo2023mechanistic}
Stolfo, A., Belinkov, Y., and Sachan, M.
\newblock A mechanistic interpretation of arithmetic reasoning in language models using causal mediation analysis.
\newblock \emph{Empirical Methods in Natural Language Processing (EMNLP)}, 2023.
\newblock URL \url{https://openreview.net/forum?id=aB3Hwh4UzP}.

\bibitem[Sun et~al.(2023)Sun, Qi, Zhang, Liu, Wang, and Huang]{sun2023tokenization}
Sun, K., Qi, P., Zhang, Y., Liu, L., Wang, W., and Huang, Z.
\newblock Tokenization consistency matters for generative models on extractive {NLP} tasks.
\newblock \emph{Empirical Methods in Natural Language Processing (EMNLP)}, 2023.
\newblock URL \url{https://aclanthology.org/2023.findings-emnlp.887}.

\bibitem[Teknium(2023)]{teknium2023bpe_weird}
Teknium.
\newblock How did the gpt tokenizer get created?, 2023.
\newblock URL \url{https://twitter.com/Teknium1/status/1634667026739527680?s=20}.

\bibitem[Touvron et~al.(2023{\natexlab{a}})Touvron, Lavril, Izacard, Martinet, Lachaux, Lacroix, Rozi{\`e}re, Goyal, Hambro, Azhar, et~al.]{touvron2023llama1}
Touvron, H., Lavril, T., Izacard, G., Martinet, X., Lachaux, M.-A., Lacroix, T., Rozi{\`e}re, B., Goyal, N., Hambro, E., Azhar, F., et~al.
\newblock Llama: Open and efficient foundation language models.
\newblock \emph{arXiv:2302.13971}, 2023{\natexlab{a}}.
\newblock URL \url{https://arxiv.org/abs/2302.13971}.

\bibitem[Touvron et~al.(2023{\natexlab{b}})Touvron, Martin, Stone, Albert, Almahairi, Babaei, Bashlykov, Batra, Bhargava, Bhosale, Bikel, Blecher, Ferrer, Chen, Cucurull, Esiobu, Fernandes, Fu, Fu, Fuller, Gao, Goswami, Goyal, Hartshorn, Hosseini, Hou, Inan, Kardas, Kerkez, Khabsa, Kloumann, Korenev, Koura, Lachaux, Lavril, Lee, Liskovich, Lu, Mao, Martinet, Mihaylov, Mishra, Molybog, Nie, Poulton, Reizenstein, Rungta, Saladi, Schelten, Silva, Smith, Subramanian, Tan, Tang, Taylor, Williams, Kuan, Xu, Yan, Zarov, Zhang, Fan, Kambadur, Narang, Rodriguez, Stojnic, Edunov, and Scialom]{touvron2023llama2}
Touvron, H., Martin, L., Stone, K., Albert, P., Almahairi, A., Babaei, Y., Bashlykov, N., Batra, S., Bhargava, P., Bhosale, S., Bikel, D., Blecher, L., Ferrer, C.~C., Chen, M., Cucurull, G., Esiobu, D., Fernandes, J., Fu, J., Fu, W., Fuller, B., Gao, C., Goswami, V., Goyal, N., Hartshorn, A., Hosseini, S., Hou, R., Inan, H., Kardas, M., Kerkez, V., Khabsa, M., Kloumann, I., Korenev, A., Koura, P.~S., Lachaux, M.-A., Lavril, T., Lee, J., Liskovich, D., Lu, Y., Mao, Y., Martinet, X., Mihaylov, T., Mishra, P., Molybog, I., Nie, Y., Poulton, A., Reizenstein, J., Rungta, R., Saladi, K., Schelten, A., Silva, R., Smith, E.~M., Subramanian, R., Tan, X.~E., Tang, B., Taylor, R., Williams, A., Kuan, J.~X., Xu, P., Yan, Z., Zarov, I., Zhang, Y., Fan, A., Kambadur, M., Narang, S., Rodriguez, A., Stojnic, R., Edunov, S., and Scialom, T.
\newblock Llama 2: Open foundation and fine-tuned chat models.
\newblock \emph{arXiv:2307.09288}, 2023{\natexlab{b}}.
\newblock URL \url{https://arxiv.org/abs/2307.09288}.

\bibitem[Wang \& Komatsuzaki(2021)Wang and Komatsuzaki]{wang2021gptj}
Wang, B. and Komatsuzaki, A.
\newblock {GPT-J-6B: A 6 Billion Parameter Autoregressive Language Model}, 2021.
\newblock URL \url{https://huggingface.co/EleutherAI/gpt-j-6b}.

\bibitem[Wei(2023)]{wei2023secondletter}
Wei, J.
\newblock Sorting a list of words by the second letter, 2023.
\newblock URL \url{https://x.com/_jasonwei/status/1661781746759909376?s=20}.

\bibitem[Wei et~al.(2022)Wei, Wang, Schuurmans, Bosma, brian ichter, Xia, Chi, Le, and Zhou]{wei2022cot}
Wei, J., Wang, X., Schuurmans, D., Bosma, M., brian ichter, Xia, F., Chi, E.~H., Le, Q.~V., and Zhou, D.
\newblock Chain of thought prompting elicits reasoning in large language models.
\newblock \emph{Neural Information Processing Systems (NeurIPS)}, 2022.
\newblock URL \url{https://openreview.net/forum?id=_VjQlMeSB_J}.

\bibitem[Yu et~al.(2023)Yu, Simig, Flaherty, Aghajanyan, Zettlemoyer, and Lewis]{yu2023megabyte}
Yu, L., Simig, D., Flaherty, C., Aghajanyan, A., Zettlemoyer, L., and Lewis, M.
\newblock {MEGABYTE}: Predicting million-byte sequences with multiscale transformers.
\newblock \emph{Neural Information Processing Systems (NeurIPS)}, 2023.
\newblock URL \url{https://openreview.net/forum?id=JTmO2V9Xpz}.

\bibitem[Zhou et~al.(2022)Zhou, Nova, Larochelle, Courville, Neyshabur, and Sedghi]{zhou2022algo_prompting}
Zhou, H., Nova, A., Larochelle, H., Courville, A., Neyshabur, B., and Sedghi, H.
\newblock Teaching algorithmic reasoning via in-context learning.
\newblock \emph{arXiv:2211.09066}, 2022.
\newblock URL \url{https://arxiv.org/abs/2211.09066}.

\bibitem[Zhou et~al.(2023)Zhou, Bradley, Littwin, Razin, Saremi, Susskind, Bengio, and Nakkiran]{zhou2023rasp}
Zhou, H., Bradley, A., Littwin, E., Razin, N., Saremi, O., Susskind, J., Bengio, S., and Nakkiran, P.
\newblock What algorithms can transformers learn? a study in length generalization.
\newblock \emph{arXiv:2310.16028}, 2023.
\newblock URL \url{https://arxiv.org/abs/2310.16028}.

\end{thebibliography}
\bibliographystyle{icml2024}

%%%%%%%%%%%%%%%%%%%%%%%%%%%%%%%%%%%%%%%%%%%%%%%%%%%%%%%%%%%%%%%%%%%%%%%%%%%%%%%
%%%%%%%%%%%%%%%%%%%%%%%%%%%%%%%%%%%%%%%%%%%%%%%%%%%%%%%%%%%%%%%%%%%%%%%%%%%%%%%
% APPENDIX
%%%%%%%%%%%%%%%%%%%%%%%%%%%%%%%%%%%%%%%%%%%%%%%%%%%%%%%%%%%%%%%%%%%%%%%%%%%%%%%
%%%%%%%%%%%%%%%%%%%%%%%%%%%%%%%%%%%%%%%%%%%%%%%%%%%%%%%%%%%%%%%%%%%%%%%%%%%%%%%
\newpage
\appendix
\onecolumn

\section{Tokenization differences between frontier LLMs}
\label{appx:llm_tokenization_schemes}

\begin{figure}[H]
    \centering
    \includegraphics[width=0.9\textwidth]{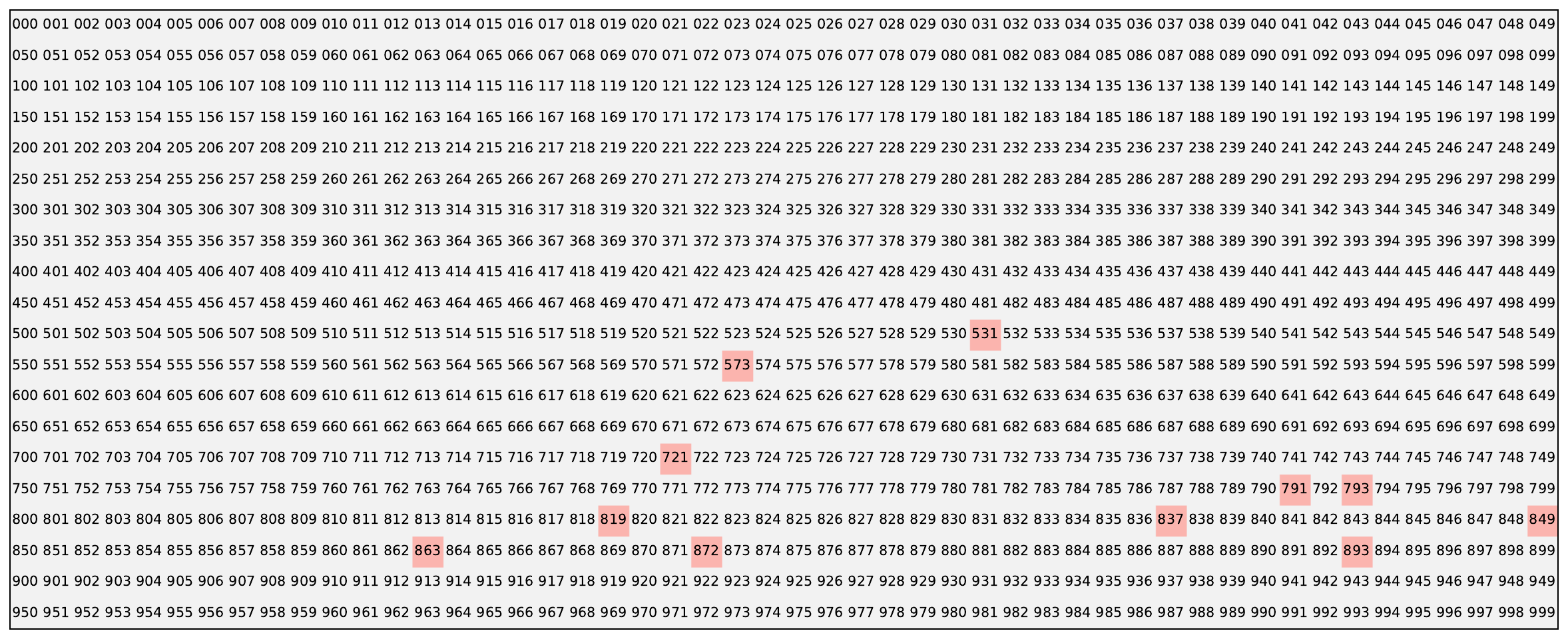}
    \caption{The equivalent of Figure~\ref{fig:p50k_tokens} but for the Claude tokenizer. All 3-digit number strings, colored red when the string does not have a corresponding single token dedicated to it. The lack of systematicity suggests that Claude tokenizes numbers using pure BPE. Note also, however, that token coverage is generally higher than in Figure~\ref{fig:p50k_tokens}, likely in part because the Claude tokenizer has a larger vocabulary size (65k tokens) than OpenAI's \texttt{p50k\_base} (50k tokens).}
    \label{fig:claude_token_coverage}
\end{figure}

\begin{figure}[H]
    \centering
    \includegraphics[width=0.4\textwidth]{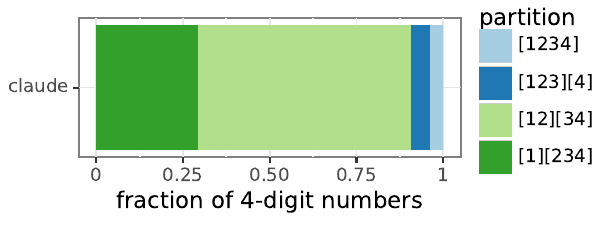}
    \caption{The equivalent of Figure~\ref{fig:4digit_tokenization_split} but for the Claude tokenizer. Note that this distribution looks more like \texttt{p50k\_base} than \texttt{cl100k\_base} in Figure~\ref{fig:4digit_tokenization_split}. This, along with Figure~\ref{fig:claude_token_coverage} above shows that Claude's tokenizer exhibits a lack of systemacity when tokenizing numbers, suggesting the use of pure BPE number tokens, rather than something bespoke (as other current models use; see Table~\ref{tab:tokenization_strategies}).}
    \label{fig:claude_4digit_partitions}
\end{figure}

\section{Experiments with other system prompts}
\label{appx:mathgpt_prompt}

\begin{figure}[H]
    \centering
    \includegraphics[width=0.5\textwidth]{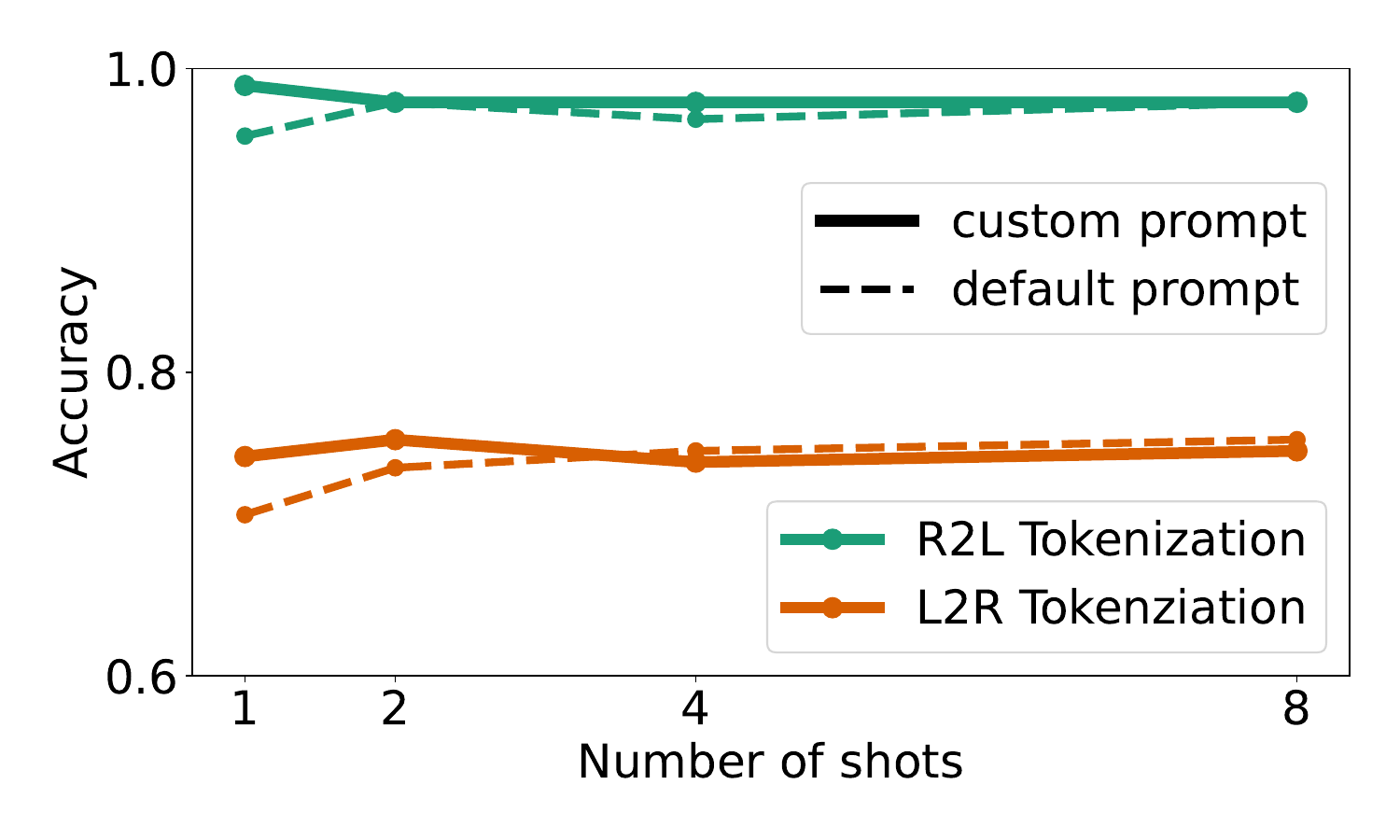}
    \caption{Comparison of R2L and L2R tokenization strategies for different numbers of shots and using two different system prompts.}
    \label{fig:mathgpt}
\end{figure}

We also conducted our main experiment with an alternate, custom system prompt (as opposed to the default \texttt{'You are a helpful assistant.'}). The prompt we used was: 

\begin{verbatim}
You are MathGPT, an expert at solving math problems. When given a math problem, 
you respond only by repeating the problem statement and appending the answer. You 
do not say any other words. 
\end{verbatim}

Results using this prompt are presented in Figure~\ref{fig:mathgpt}. We found it lead to small improvements in performance at low shot numbers (e.g., 1-shot) but these diminished at 8-shots. To maximize the reproducibility and applicability of our results, we decided to just use the default prompt. As we report 8-shot results throughout most of the paper, we doubt the system prompt would have a large effect on our results, given Figure~\ref{fig:mathgpt}.

\section{Frequency effects}
\label{appx:frequency}

Given the findings of prior work on numerical reasoning demonstrating frequency effects \cite{razeghi2022frequencies}, we also investigated whether or not our observed error patterns could be explained by term frequency. While we do not have access to the pre-training data of GPT-3.5 and GPT-4 models, we use the tokenizer merge ranks\footnote{Recall that tokens are created roughly in order of decreasing frequency in the corpus used to train the tokenizer.} as a signal of term frequency. We analyze the expanded set of problems used for the error analysis in Section~\ref{sec:error_patterns_overall}. Our results are summarized below:

\textbf{When making an error, GPT-3.5 is slightly more likely to output a more frequent token.} For each token in the model response on problems where it makes a mistake, we consider if the outputted incorrect token is more or less frequent (has lower or higher merge rank) than the correct one. Of the 25 errors made by the model when using R2L tokenization, 15 involve substituting in a more frequent token (60\%, $p=0.115$ using a binomial null distribution assuming chance is 50\%). Of the 425 errors made when using L2R tokenization, 238 involve substituting in a more frequent token (56\%, $p=0.005$ using a binomial null distribution assuming chance is 50\%). While we do see a significant effect in the L2R tokenization case, the margin is relatively small, which suggests that token frequency is not the dominant reason behind the error patterns.

\begin{figure}
    \centering
    \includegraphics[width=0.6\textwidth]{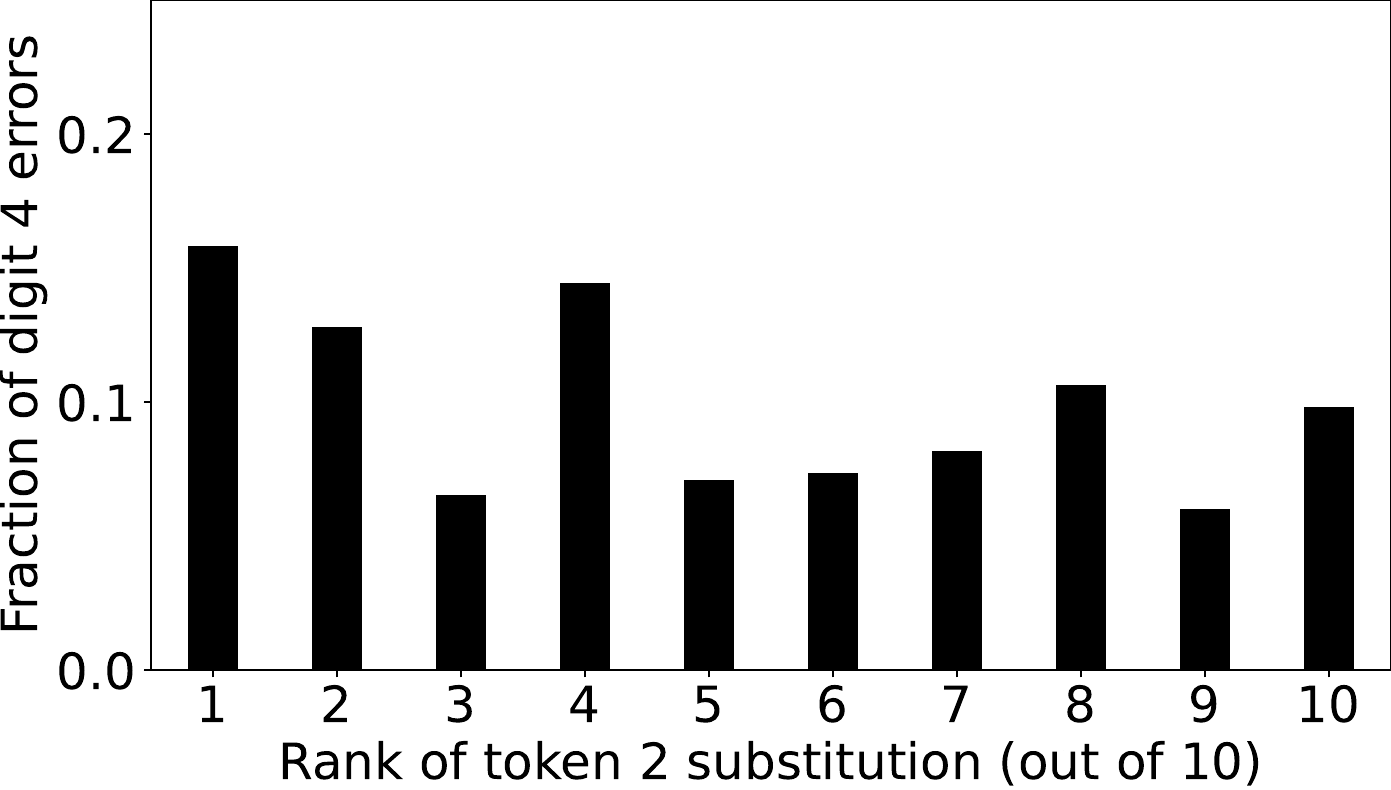}
    \caption{Distribution of relative rank of substituted incorrect token 2 in model response when using L2R tokenization in the length mismatch condition.}
    \label{fig:digit4_tok_freq}
\end{figure}

\textbf{When using L2R tokenization in the length mismatch case, GPT-3.5 errors do not show strong correlation to token frequency.} In Section~\ref{sec:error_digit4}, we found that GPT-3.5 always gets the fourth digit wrong (Figure~\ref{fig:error_digit4}a). We then found correlation in the specific error in digit 4 to the magnitude difference between correct digit and digit in the model response (Figure~\ref{fig:error_digit4}b). Here, we ask if the substituted token 2 (whose first digit would be digit 4 of the response) is correlated to frequency in training data. Specifically, for each problem, we rank the tokens corresponding to the 10 possible ``digit 4 mistakes'' by merge rank. In Figure~\ref{fig:digit4_tok_freq}, we show the distribution of ranks across all 365 problems where the model makes ``digit 4'' errors. If models are preferentially substituting in more frequent tokens, we would expect to see a negative trend from the top left to the bottom right (as we did in  Figure~\ref{fig:error_digit4}b). In Figure~\ref{fig:digit4_tok_freq}, we see a slight preference for outputting the most likely token (roughly 16\% of the time, where chance would be 10\%), but overall we see no clear trend.

\textbf{Off-by-one errors do not seem to be correlated to answer token frequency.} In Section~\ref{sec:errors_rest}, we found that the vast majority of remaining errors (for R2L tokenization, and for L2R tokenization in the length match condition) are off-by-one errors in the units digit of a token. Here, we ask if the specific substitution by the model is correlated to token frequency, measured by merge rank. Specifically, we condition on the model possibly making an off-by-one error, which means there are 3 possible output tokens (the correct token, the correct token minus one, the correct token plus one). We then rank these tokens based on merge rank, and see if the model preferentially picks the token with lowest merge rank. Of the 24 off-by-one errors when using R2L tokenization, we find the model only picks the ``most frequent'' token 7 times. Of the 53 off-by-one errors when using L2R tokenization, we find the model only picks the ``most frequent'' token 17 times. Both of these are essentially what we would expect by chance (one out of three), which suggests that output token frequency effects are not a dominant factor in why the model makes off-by-one errors.

Overall, we find mild to no evidence of token frequency effects in our experiments. This could be due to the presumably larger scale of GPT-3.5 (as compared to GPT-J, used by \citet{razeghi2022frequencies}). However, we note that our method of measuring token frequency is imperfect---relying on BPE merge ranks to signal frequency as we do not have access to pre-training data. Future work could study such associations further in newer, larger models with open pretraining data \cite{groeneveld2024olmo}.

\section{Stereotyped patterns in model log probabilities}
\label{appx:logprobs}

Mirroring the results of Section~\ref{sec:error_patterns_overall}, we found stereotyped patterns in model log probabilities (``logprob''). Specifically, the OpenAI API returns the top 5 tokens at each position with their corresponding logprobs. We analyzed these log probabilities in three cases: L2R tokenization on length mismatch problems, L2R tokenization on length match problems, and R2L tokenization on all problems. These conditions mirror the most salient error effects we found in Section~\ref{sec:error_patterns_overall}, with the former leading to ``digit 4'' errors, and the latter two leading to mostly off-by-one errors.

In addition to the raw logprobs, we computed an additional entropy metric (per output token) to measure model uncertainty in its output. Since access is restricted to the top 5 logprobs, we use the following lower bound, $H_{\text{lower}}$, to the true entropy:
\begin{align*}
H_{\text{true}} &\equiv - \sum_{i=1}^V p_i \log\left(p_i\right) \\
&= - \left(\sum_{i=1}^5 p_i \log\left(p_i\right) + \sum_{i=6}^V p_i \log\left(p_i\right) \right) \\
&\geq - \left(\sum_{i=1}^5 p_i \log\left(p_i\right) + \sum_{i=6}^V p_i \log\left(p_5\right) \right) \\
&= - \left(\sum_{i=1}^5 p_i \log\left(p_i\right) + \left(1-\sum_{i=1}^5 p_i\right) \cdot \log\left(p_5\right)\right) \\
&\equiv H_{\text{lower}},
\end{align*}
where $p_i$ denotes the probability of the $i$-th most likely token. We use the natural logarithm for entropy, so all entropies are in nats (not bits).

For the ``digit 4'' error pattern (Section~\ref{sec:error_digit4}), we find an interesting trend in model entropy. The entropy both on problems it gets incorrect (91.25\%) and correct (8.25\%) is roughly the same (2.066 and 2.061 respectively). Even when the model gets the question right, it's unsure of its answer, suggesting that it might just be guessing a second output token with the right tens and ones digit and random hundreds digit. Providing further evidence for this mechanism, we observe that, of the problems where the model makes an error, about half (49.6\%) of the time the correct answer appears in the top 5 output tokens. This is in line with what we would see for random guessing from the 10 tokens. That said, the model may exhibit some degree of bias towards the correct output, as evidenced by the downward trend in Figure~\ref{fig:error_digit4}b.

\begin{figure}
    \centering
    \includegraphics[width=0.6\textwidth]{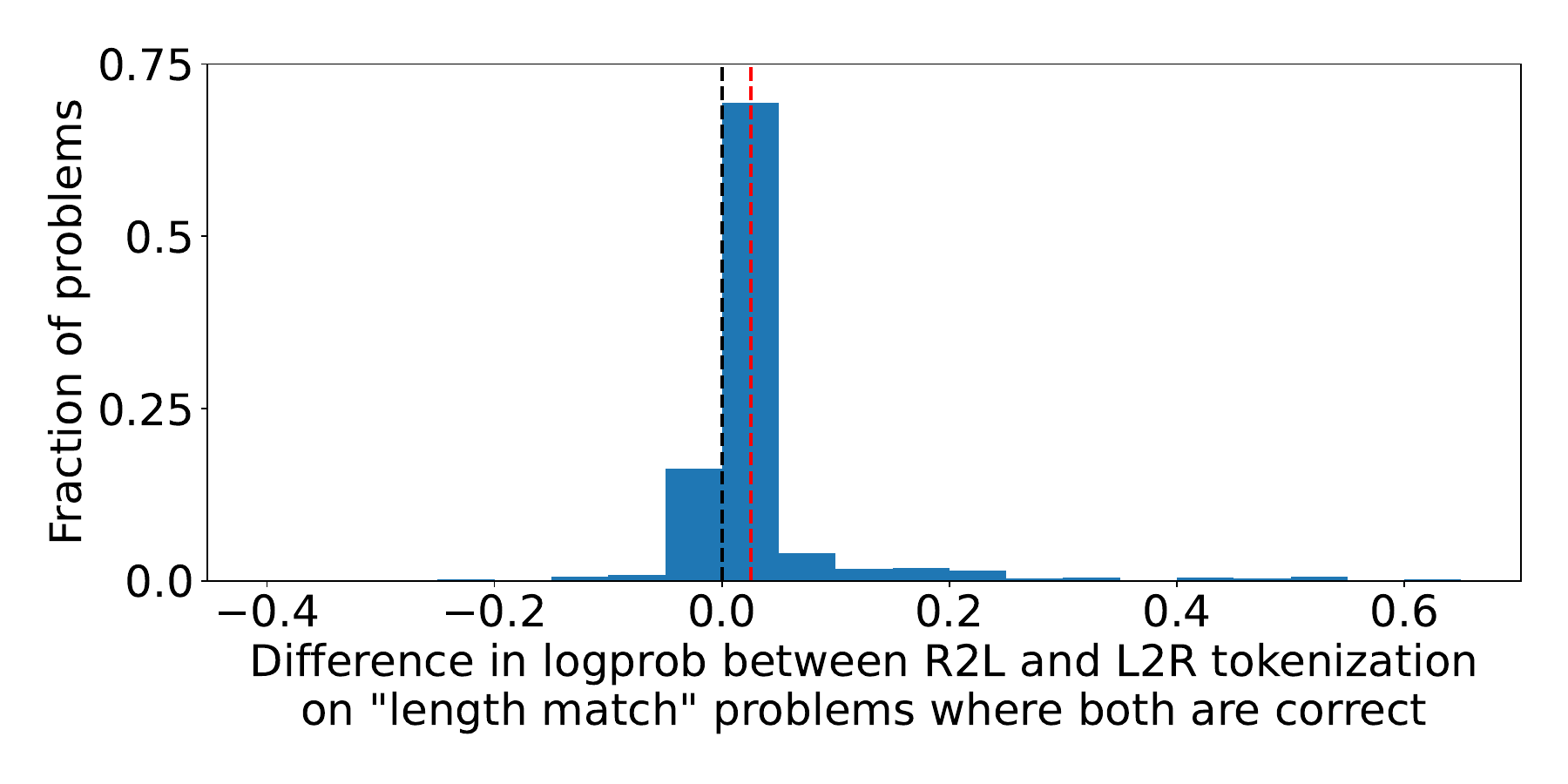}
    \vspace{-2em}
    \caption{Histogram of difference in answer log probabilities (summed over tokens) on length match problems that the model answers correctly using both R2L and L2R tokenization. Black dotted line signifies 0. Red dotted line shows the average difference---on average, the model is more ``confident'' when using R2L tokenization.}
    \label{fig:correct_logprob_hist}
\end{figure}

For the off-by-one error patterns (Section~\ref{sec:errors_rest}), we observe a qualitatively different trend. Specifically, of the 53 off-by-one errors when using L2R tokenization on length match problems, in all cases we find that the second most likely token is the correct answer. We observe the same effect on the 25 off-by-one errors when using R2L tokenization. Furthermore, the entropy in both cases is around $0.45 \pm 0.05$, indicating that the model puts most of its weight on these top 2 most likely tokens. Unlike in the ``digit 4'' case, model entropy on correct problems is significantly lower (approximately 0.03, averaged across dataset and tokens) indicating that the model is ``confidently correct'' when using L2R tokenization on length match problems or R2L tokenization on all problems. Interestingly, on the subset of length match problems that the model answers correctly in both L2R and R2L tokenization, we found the model is slightly more confident when using R2L tokenization (which aligns with our intuition, as the model is also more often correct when using R2L tokenization)---see Figure~\ref{fig:correct_logprob_hist}.

These results demonstrate that, depending on tokenization direction and alignment between input and output tokenization, we observe stereotyped patterns in model log probabilities. When using L2R tokenization on length mismatch problems, the model appears to make a magnitude-biased guess between all possible fourth digits (corresponding to 10 possible tokens\footnote{A completely random guess over 10 tokens would correspond to an entropy of about 2.3, which is in line with the lower bound observed (of about 2.06) and the finding of a slight mangitude bias (which would decrease the entropy from 2.3).}). In the other cases, the model is mostly confidently correct. When it does make an error, it's almost always an off-by-one error (Section~\ref{sec:errors_rest}) where it's uncertain between its chosen off-by-one incorrect answer and the true answer, but does not really consider other outputs beyond these two.\footnote{A completely random guess over 2 tokens would correspond to an entropy of about 0.69, which is in line with the lower bound observed (of about 0.45).}

\section{Additional experimental details}
\label{appx:extra_details}

All code and raw results can be found at \url{https://github.com/aadityasingh/TokenizationCounts}.

\subsection{Length control for error analysis}
\label{appx:length_control}

As described in Section~\ref{sec:error_length}, after noticing errors mostly come from the length mismatch condition in our original experiments (which used 90 problems, balanced by input digit length), we conducted a larger experiment where we controlled for input and output digit lengths. Specifically, we considered the following (addend1\_length, addend2\_length, answer\_length) triplets: (7,7,7), (7,7,8), (8,7,8), (7,8,8), (8,7,9), (7,8,9), (8,8,8), (8,8,9), (9,7,9), (7,9,9), (9,8,9), (8,9,9), (9,9,9). Problems in each condition were sampled randomly so as to satisfy the digit length constraints for each triplet. We sampled 100 problems for each triplet, for a total of 1300 problems.

\subsection{Access dates}
\label{appx:access}

Given the changing nature of the OpenAI API, we report access dates for all experiments below. We tried to use the supposed ``fixed'' models for all experiments, but did notice some non-determinism, even at temperature 0---an issue that may be due to non-determinism in floating point arithmetic. We also note that the \texttt{gpt-4-0314} appears to have been early-deprecated, as we can no longer access it despite the supposed June 13, 2024 deprecation date on \url{https://platform.openai.com/docs/deprecations}.

Access dates by figure in main text:
\begin{itemize}
    \item \texttt{gpt-3.5-turbo-0301}, Figure~\ref{fig:token_dir_shots}: April 7, 2023
    \item \texttt{gpt-3.5-turbo-0301}, Figure~\ref{fig:control_delim}: May 18, 2023
    \item \texttt{gpt-3.5-turbo-0301}, Figure~\ref{fig:control_thinking_token} left two columns: May 18, 2023
    \item \texttt{gpt-3.5-turbo-0301}, Figure~\ref{fig:control_thinking_token} right two columns: April 7, 2023
    \item \texttt{gpt-3.5-turbo-0301}, Figure~\ref{fig:answer_length}-\ref{fig:carries}: January 25, 2024
    \item \texttt{gpt-3.5-turbo-0301}, Figure~\ref{fig:repeat_style}-\ref{fig:answer_style} May 24, 2024
    \item \texttt{gpt-4-0314}, Figure~\ref{fig:main_effect_new_models}: May 2, 2023
    \item \texttt{gpt-3.5-turbo-0613}, Figure~\ref{fig:main_effect_new_models}-\ref{fig:error_new_models}: January 25, 2024
     \item \texttt{gpt-3.5-turbo-1106}, Figure~\ref{fig:main_effect_new_models}-\ref{fig:error_new_models}: January 29, 2024
    \item \texttt{gpt-4-0613}, Figure~\ref{fig:main_effect_new_models}-\ref{fig:error_new_models}: January 25, 2024
    \item \texttt{gpt-4-1106-preview}, Figure~\ref{fig:main_effect_new_models}-\ref{fig:error_new_models}: January 29, 2024
\end{itemize}

\subsection{Example prompts}
\label{appx:prompts}

In this section, we provide example prompts we used for various experiments. For simplicity, we use the same query for each prompt shown below, and we only use 2 shots (most experiments in the main text are done with 8 shots). In practice, we sampled shots randomly (controlling for digit length to match the query length) for each query, as explained in Section~\ref{sec:experiment_setup}. For the experiments described in Section~\ref{sec:error_patterns_overall} and Appendix~\ref{appx:length_control}, the shots were also controlled to have the same answer length as the query. The examples we present below, though, are for the runs in the rest of the paper (where only input digit lengths are controlled). For maximum clarity, we display prompts as the list of dictionaries that gets sent to OpenAI's API and roughly in the order used for figures in the paper. Following the advice at \texttt{https://platform.openai.com/docs/guides/prompt-engineering/tactic-provide-examples}, we make use of the multi-turn chat dialog to prompt the model, as opposed to one big user message with all the examples.

% Use figure environments to make sure description + prompt are all on the same page.

\begin{figure}[H]
L2R tokenization, input-digit-controlled for two 7-digit numbers:
    \begin{verbatim}
[{'role': 'system', 'content': 'You are a helpful assistant.'}, 
{'role': 'user', 'content': '3790206+6739555='}, 
{'role': 'assistant', 'content': '10529761'}, 
{'role': 'user', 'content': '6777159+7096168='}, 
{'role': 'assistant', 'content': '13873327'}, 
{'role': 'user', 'content': '8302080+3529456='}]
\end{verbatim}
\end{figure}

\begin{figure}[H]
R2L tokenization, input-digit-controlled for two 7-digit numbers:
    \begin{verbatim}
[{'role': 'system', 'content': 'You are a helpful assistant.'}, 
{'role': 'user', 'content': '3,790,206+6,739,555='}, 
{'role': 'assistant', 'content': '10,529,761'}, 
{'role': 'user', 'content': '6,777,159+7,096,168='}, 
{'role': 'assistant', 'content': '13,873,327'}, 
{'role': 'user', 'content': '8,302,080+3,529,456='}]
\end{verbatim}
\end{figure}

\begin{figure}[H]
R2L tokenization, delimiter-control condition using \texttt{'\#'}:
    \begin{verbatim}
[{'role': 'system', 'content': 'You are a helpful assistant.'}, 
{'role': 'user', 'content': '3#790#206+6#739#555='}, 
{'role': 'assistant', 'content': '10#529#761'}, 
{'role': 'user', 'content': '6#777#159+7#096#168='}, 
{'role': 'assistant', 'content': '13#873#327'}, 
{'role': 'user', 'content': '8#302#080+3#529#456='}]
\end{verbatim}
\end{figure}

\begin{figure}[H]
L2R tokenization, thinking token control by using separators in L2R tokenization:
    \begin{verbatim}
[{'role': 'system', 'content': 'You are a helpful assistant.'}, 
{'role': 'user', 'content': '379,020,6+673,955,5='}, 
{'role': 'assistant', 'content': '105,297,61'}, 
{'role': 'user', 'content': '677,715,9+709,616,8='}, 
{'role': 'assistant', 'content': '138,733,27'}, 
{'role': 'user', 'content': '830,208,0+352,945,6='}]
\end{verbatim}
\end{figure}

\begin{figure}[H]
L2R tokenization, thinking token control by using 2 extra spaces:
    \begin{verbatim}
[{'role': 'system', 'content': 'You are a helpful assistant.'}, 
{'role': 'user', 'content': '3790206  +  6739555  =  '}, 
{'role': 'assistant', 'content': '10529761'}, 
{'role': 'user', 'content': '6777159  +  7096168  =  '}, 
{'role': 'assistant', 'content': '13873327'}, 
{'role': 'user', 'content': '8302080  +  3529456  =  '}]
\end{verbatim}
\end{figure}

\begin{figure}[H]
L2R tokenization, thinking token control by using 2 extra spaces:
    \begin{verbatim}
[{'role': 'system', 'content': 'You are a helpful assistant.'}, 
{'role': 'user', 'content': '3790206  +  6739555  =  '}, 
{'role': 'assistant', 'content': '10529761'}, 
{'role': 'user', 'content': '6777159  +  7096168  =  '}, 
{'role': 'assistant', 'content': '13873327'}, 
{'role': 'user', 'content': '8302080  +  3529456  =  '}]
\end{verbatim}
\end{figure}

\begin{figure}[H]
Repeat L2R $\rightarrow$ R2L:
    \begin{verbatim}
[{'role': 'system', 'content': 'You are a helpful assistant.'}, 
{'role': 'user', 'content': '3790206+6739555='}, 
{'role': 'assistant', 'content': '3,790,206+6,739,555=10,529,761'}, 
{'role': 'user', 'content': '6777159+7096168='}, 
{'role': 'assistant', 'content': '6,777,159+7,096,168=13,873,327'}, 
{'role': 'user', 'content': '8302080+3529456='}]
\end{verbatim}
\end{figure}

\begin{figure}[H]
Repeat control L2R $\rightarrow$ L2R:
    \begin{verbatim}
[{'role': 'system', 'content': 'You are a helpful assistant.'}, 
{'role': 'user', 'content': '3790206+6739555='}, 
{'role': 'assistant', 'content': '3790206+6739555=10529761'}, 
{'role': 'user', 'content': '6777159+7096168='}, 
{'role': 'assistant', 'content': '6777159+7096168=13873327'}, 
{'role': 'user', 'content': '8302080+3529456='}]
\end{verbatim}
\end{figure}

\begin{figure}[H]
Output control L2R $\rightarrow$ R2L:
    \begin{verbatim}
[{'role': 'system', 'content': 'You are a helpful assistant.'}, 
{'role': 'user', 'content': '3790206+6739555='}, 
{'role': 'assistant', 'content': '10,529,761'}, 
{'role': 'user', 'content': '6777159+7096168='}, 
{'role': 'assistant', 'content': '13,873,327'}, 
{'role': 'user', 'content': '8302080+3529456='}]
\end{verbatim}
\end{figure}

\end{document}